\documentclass[10pt,twocolumn,letterpaper]{article}

\usepackage[pagenumbers]{wacv} 

\usepackage{lineno}
\usepackage{amsmath}
\usepackage{amssymb}
\usepackage{mathtools}
\usepackage{makecell}
\usepackage{arydshln}
\usepackage{url}
\usepackage{booktabs}       
\usepackage{amsfonts}       
\usepackage{amsmath}
\usepackage{xcolor}         
\usepackage[numbers]{natbib}
\usepackage{multibib} 
\newcites{supp}{References} 
\MakeRobust{\citesupp} 
\usepackage{sidecap}
\usepackage{graphicx}
\usepackage{caption}
\usepackage{ulem}

\usepackage{amsmath}
\usepackage{comment}
\usepackage{multirow,bigdelim}
\usepackage{lipsum}
\usepackage{array}
\usepackage{wrapfig}

\usepackage{adjustbox}
\usepackage[percent]{overpic}
\usepackage{makecell}
\usepackage{xcolor}
\usepackage{soul}
\usepackage{amsfonts}
\usepackage{xspace}
\usepackage{amsmath}

\makeatletter
\newcommand{\settitle}{\@maketitle}
\makeatother

\newcolumntype{C}[1]{>{\centering\let\newline\\\arraybackslash\hspace{0pt}}m{#1}}

\newif\ifdraft
\drafttrue

\ifdraft
\definecolor{darkpink}{rgb}{0.561, 0.282, 0.427}

\newcommand{\abc}[1]{{\color{purple}[\textbf{AB:} #1]}}

\newcommand{\drop}[1]{}

\else
\newcommand{\dcc}[1]{}
\newcommand{\rgc}[1]{}
\newcommand{\opc}[1]{}
\newcommand{\gcc}[1]{}
\newcommand{\hmc}[1]{}
\newcommand{\abc}[1]{}

\fi

\newcommand{\ourmethod}{\textbf{ElasticDino}\xspace}

\makeatother

\usepackage[bottom]{footmisc}
\usepackage{arydshln}

\makeatletter
\def\blfootnote{\xdef\@thefnmark{}\@footnotetext}
\makeatother

\usepackage{caption}

\usepackage{siunitx}    
\definecolor{wacvblue}{rgb}{0.21,0.49,0.74}
\usepackage[pagebackref,breaklinks,colorlinks,allcolors=wacvblue]{hyperref}

\def\wacvPaperID{1102} 
\def\confName{WACV}
\def\confYear{2027}

\title{\ourmethod: Dense Semantic Features using Elastic Deformation}

\author{Ulysse Mizrahi$^*$\hspace{4mm}
Sean Zaretzky$^*$\hspace{4mm}
Dov Danon\hspace{4mm}
Amit H. Bermano\\[4pt]
Tel Aviv University\\
\small\url{https://odusseys.github.io/ElasticDino/}\\
}

\begin{document}
\makeatletter
\let\SavedMaketitle\maketitle
\let\Saved@maketitle\@maketitle
\makeatother
\maketitle
{\renewcommand{\thefootnote}{*}\footnotetext{Equal contribution.}}
\begin{figure}[ht]
  \centering
    \includegraphics[trim={0cm 0 25.5cm 0cm},clip, width=.99\linewidth]{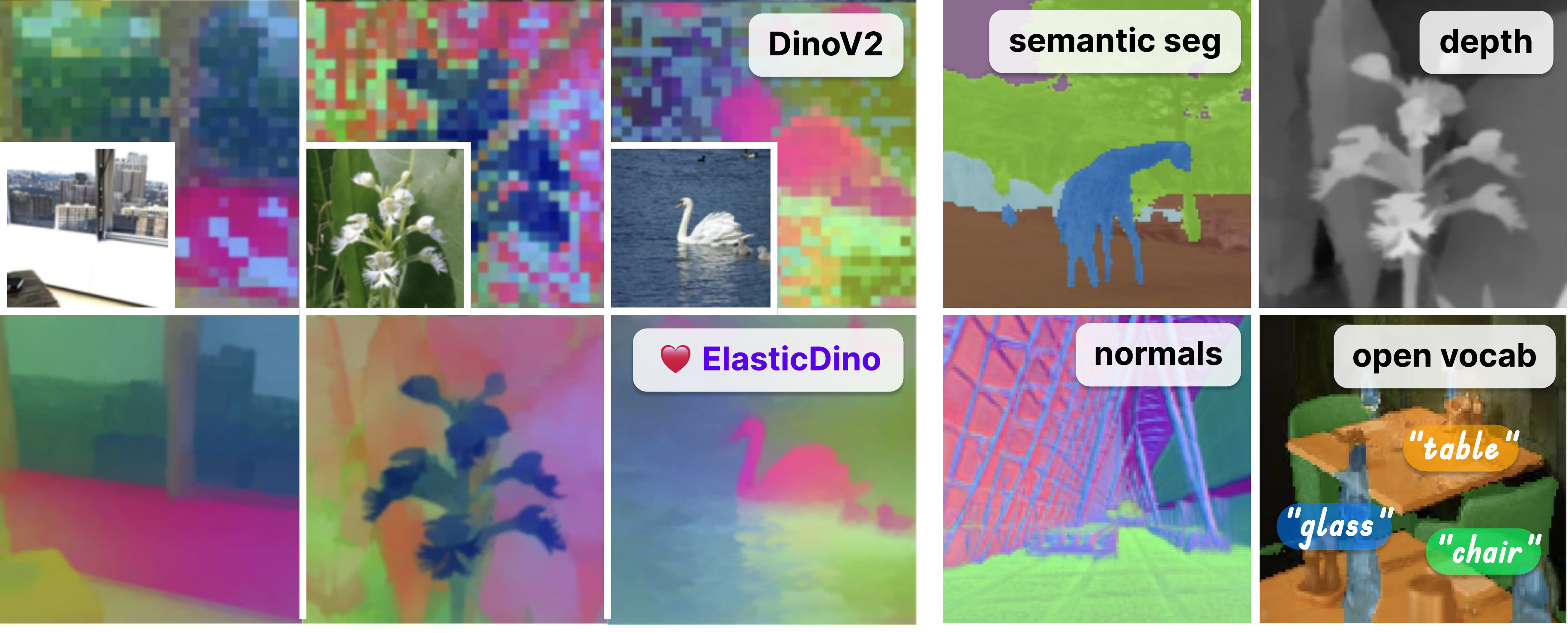}
    \includegraphics[trim={38cm 12.8cm 0cm 0cm},clip,width=.49\linewidth]{teaser.pdf}
    \includegraphics[trim={38cm 0.25cm 0cm 12.6cm},clip,width=.49\linewidth]{teaser.pdf}
   \caption{
   ElasticDino upscales ViT features and corrects their artifacts, allowing to efficiently train a variety of dense vision tasks.
   }
   \label{fig:teaser}
\end{figure}

\begin{abstract}

Vision transformers (ViTs) serve as the backbone for countless vision tasks, including dense prediction tasks, such as segmentation and depth estimation, even though the quadratic complexity of self-attention severely limits feature resolution and induces spatial artifacts. In this work, we introduce \ourmethod, a lightweight adapter which produces high-resolution feature maps, adhering to the fine details of the input image. As a key insight, we observe the consistency of artifacts across channels, and simultaneously resolve them to match the input image structure. Using a slim architecture based on spatial deformations, \ourmethod produces pixel-aligned, high-resolution feature maps efficiently. Leveraging deformations allows strictly preserving sharp semantics, and thereby avoid the smoothness biases inherent to alternative approaches. Due to the absence of ground truth, we employ pixel-level linear heads to predict a set of downstream dense tasks, guiding the deformation process without leaking task specific semantics into the model. We demonstrate superior accuracy over current methods on a range of popular dense task benchmarks, with better computational performance.

\end{abstract}

\section{Introduction}\label{sec:intro}


Foundational Visual features have become the backbone for most vision tasks in recent literature. Under this paradigm, a large foundational model is typically contrastively trained to embed images into a latent space. This embedding is rich and semantic by nature, and serves as a great starting point for various downstream tasks with significantly less training requirements. Specifically, self-supervised features trained using a Vision Transformer (ViT) architecture (i.e., DINO \citep{caron2021emerging,oquab2023dinov2,simeoni2025dinov3}) are employed in state-of-the-Art semantic, panoptic and instance segmentation \citep{cheng2021mask2former,jain2022oneformertransformerruleuniversal}, object detection \citep{liu2024groundingdinomarryingdino}, monocular depth estimation \citep{ranftl2020robustmonoculardepthestimation,yang2024depthanythingunleashingpower, yang2024depthv2, bochkovskii2025depthprosharpmonocular}, optical flow estimation \citep{tumanyan2024dinotrackertamingdinoselfsupervised}, and more.

Despite their stellar success, these models have an inherent flaw: built on self-attention, they require quadratic memory, imposing patchification of the input image, effectively restricting the resolution of the final feature maps and introducing artifacts in their semantics. Indeed, for most hardware, a patched resolution of $32\times32$ or $64\times64$ is typically used, even when training dense tasks such as segmentation or depth estimation. Typically, a densification head is added to recover full resolution, (e.g. DPT \citep{ranftl2021visiontransformersdenseprediction}), which requires large amounts of data to recover fine details, as attested by the large and complex datasets used to train monocular depth estimation tasks \citep{yang2024depthanythingunleashingpower, yang2024depthv2}.
This approach implicitly also requires training a specific head for each downstream task due to poor transferability.

A different approach that has recently emerged in the literature \cite{fu2024featup,huang2025loftup,couairon2025jafar,wimmer2025anyup, walmer2026uplift} seeks to enhance the resolution of the produced features, indifferent the downstream task. As we demonstrate, such approaches still suffer from performance issues due to attention-based layers, and base their upscaling process on reconciling high and low resolution features through cropping-like mechanisms, leading to inevitable bleeding of feature semantics at boundaries. 

In this paper, we introduce \textit{\ourmethod}, a lightweight upscaling method to enhance DINO features, yielding efficient and sharp result through a deformation-based architecture and simple training through a collection of minimalistic heads. 

To achieve this we first observe that  artifacts are consistent across feature-channels, and can thus be resolved for all channels simultaneously (see Figure~\ref{fig:simultaneous}, subsection~\ref{subsec:cross_channel_consistency}). As such, matching of feature and image structure our primary goal, regardless of specific per-chanel values. Second, we observe that most feature artifacts stem from the added positional encoding or under-sampling of semantic edges, resulting in thickening of objects and region bleeding. Hence, we learn deformation fields that align all feature channels to the image structure, alternating between upsampling and deformation. We observe that deformations offer efficient structure  preservation, inherently and consistently adhering to sharp semantics. This alleviates otherwise inherent smoothness biases that are typical to the mechanisms used by previous work, and hence minimizes bleeding and fuzzy transition artifacts. Critically, since no upscaled ground truth exists in our setting, instead of cropping or further under-sampling approaches, we train \ourmethod using a set of five lightweight downstream heads, each learning to predict a different dense map, while limited to a $1x1$ receptive field, forcing pixel-level matching of feature semantics.

Through careful evaluation, we demonstrate that \ourmethod produces high-quality alignment of semantics and structure, while removing artifacts such as those induced by positional encoding. Using lightweight heads, we report superior results compared to other upscaling methods on semantic segmentation, open-vocabulary segmentation, and depth estimation, with minimal computational effort. 



\section{Related work}\label{sec:related}

\subsection{Vision transformers and semantic features}



Vision transformers have shown exceptional performance on vision tasks, being able to model complex long-range dependencies, and setting state-of-the-art results in a variety of vision tasks such as classification \citep{zhai2022scalingvisiontransformers}, segmentation \citep{cheng2021mask2former, jain2022oneformertransformerruleuniversal}, object detection \citep{liu2024groundingdinomarryingdino}, monocular depth estimation, \citep{yang2024depthanythingunleashingpower, bochkovskii2025depthprosharpmonocular, yang2024depthv2}, joint text-image encoding \citep{radford2021learningtransferablevisualmodels}, Vision Large Language Models \citep{wang2022simvlmsimplevisuallanguage}, and image generation \citep{peebles2023scalablediffusionmodelstransformers}.

Many of the above tasks relied on pre-trained ViT encoders, much like vision models previously relied on convolutional backbones such as ResNet \citep{he2015deepresiduallearningimage}. Indeed, ViTs have shown the ability to generate spatial semantic features, which transfer to a variety of tasks. Chief among these is the DINO family \citep{caron2021emerging,oquab2023dinov2,simeoni2025dinov3}, which through unsupervised training provides exceptionally strong semantic features, and has since become the go-to backbone for various state-of-the-art models \citep{yang2024depthanythingunleashingpower,liu2024groundingdinomarryingdino}.

\subsection{Correction of ViT artifacts}

In order to correct for artifacts in ViT (in particular DINO) activations, several methods have been introduced - though without considering higher feature resolutions.

Soon after the introduction of DINO, it was realized that a large part of the artifacts encountered were due to a leakage of global semantics in patch features \citep{darcet2024visiontransformersneedregisters}. "Register" tokens were introduced and added to the tokenized image sequence, meant to capture global semantics, while letting patch features absorb relatively more local semantics. 

\textbf{Self-Patch} \citep{yun2022patchlevelrepresentationlearningselfsupervised} improves upon DINO's self-supervision approach by enforcing local invariances among neighboring patches, improving spatial consistency and downstream performance although without upsampling features to higher resolutions. This however requires retraining of the model.

More recently, \textbf{Denoising Vision Transformers} \citep{yang2024denoisingvisiontransformers} was introduced to mainly resolve the artifacts introduced by positional encodings. They achieve this through decomposing feature maps into a "ground truth" component and several shared positional noise components. They impose consistency between the features produced by image crops, thus removing positional information from the image features.
It is important to note that positional information in feature maps is nonetheless important; removing it entirely severely reduces the out-of-the-box ability to model positional dependencies between objects or points of the image.

\subsection{ViTs and dense vision tasks}

The first ViT densification approaches were task specific, with \textbf{Dense Prediction Transformer (DPT)} \citep{ranftl2021visiontransformersdenseprediction} still being a popular architecture. It combines a ViT backbone with a convolutional decoder to generate high-resolution depth maps, leading to success in tasks such as monocular depth estimation \citep{ranftl2020robustmonoculardepthestimation,yang2024depthanythingunleashingpower,yang2024depthv2}. In addition, \textbf{ViT-Adapter} \citep{chen2023visiontransformeradapterdense} adds a trainable module to a ViT, injecting vision-specific priors. DySample \cite{liu2023learning} dynamic sampling, SAPA \cite{lu2022sapa} and ReSFU \cite{zhou2025refreshed} further refine spatial semantics. While displaying impressive performance, these approaches struggle with high-frequency details, and more importantly demand data-intensive per-task training. Thus inducing high costs and restricting applicability to specialized domains.

More recently, a series of task agnostic literature has emerged, surpassing results of the task specific approaches while offering better generalization.  \textbf{FeatUp} \citep{fu2024featup} introduces two separate upscaling methods, both based on reconciling features for various augmentations of an image. The first is trained per-image and demonstrates honorable performance, but requires over a minute of training for every image, precluding most real-world applications. The second method results in a generic upscaler, which displays gross artifacts - unavoidable with the joint-bilataral-upsampling kernels employed.

 Similarly, \textbf{Lift} \citep{suri2024liftsurprisinglysimplelightweight} uses a simple CNN architecture which combines image and features and then upsamples them. The training procedure consists in matching upscaled features produced at a lower image resolution to features produced at higher image resolution, and not upscaled. As the chosen "ground truth" is not free of artifacts, the incentive to fully resolve them is not complete. Further, the editing of feature values, which are mixed with image encodings, does not guarantee strict preservation of semantics.

\textbf{LoftUp} \citep{huang2025loftup} passes images, features and positional encodings to cross-attention blocks. The training procedure is two-staged. First, they compute various segmentation masks on the image, use a form of bilateral upsampling guided by the masks, and use these as pseudo-labels. In a second stage, they perform teacher-student distillation, now using as objective the reconciliation of feature values along crops. This suffers the same issues, with direct feature value editing, and reduced incentive to correct artifacts, especially bleeding.

\textbf{JAFAR} \citep{couairon2025jafar} uses cross-attention using the encoded image as the query, features as value, and both as key, to compute pixel-to-semantics attention scores and query them within the feature values. This ensures that feature semantics are preserved. The training procedure is identical to LIFT and as such lacks the incentive to fully resolve artifacts.

\textbf{AnyUp} \citep{wimmer2025anyup} proposes a backbone-agnostic upsampling module capable of transferring across models without retraining the upscaling head. To address the computational inefficiencies and the tendency of global attention to reference distant, unrelated image areas, AnyUp restricts its cross-attention computation to local spatial windows, supplemented by a feature-agnostic layer to enable transferability. However, the training procedure relies on matching features from randomly sampled image crops. Similar to FeatUp, this formulation does not provide a strong enough incentive to fully resolve the inherent spatial artifacts of the backbone.

\textbf{UPLiFT} \citep{walmer2026uplift} addresses the heavy memory bottlenecks of single-step cross-attention approaches by proposing an iterative upsampling framework. It relies on an encoder-decoder architecture equipped with a local attention pooling operator to maintain linear complexity with respect to the number of visual tokens. Similar to LIFT, the training procedure optimizes a multi-step reconstruction loss that matches iteratively upscaled features from low-resolution inputs to the standard features extracted from high-resolution inputs. By explicitly treating the raw backbone features as the ground truth target, this formulation inherits the backbone's spatial artifacts and provides no incentive to fully resolve them.
\section{Method}\label{sec:method}

\ourmethod is an upscaling network for DINO features, learned as iterative upsampling followed by elastic deformations of the features. Given an image and the corresponding feature maps generated by the ViT, our goal is to upscale the feature maps while matching the image structure. Figure \ref{fig:architecture} depicts an overview of the architecture: The input features are repeatedly upscaled and deformed, with the deformation field being predicted using convolutional layers. We note that the advantages of this approach are lightweight computation that allow for higher-resolution maps at negligible cost with respect to backbone evaluation, and better handling of high-frequency edges. Another major advantage of this approach is that since we sample from the original feature map, we diminish hallucination hazards, ensuring strong structural and semantic preservation. 

As there are no ground truth upscaled features available, we train $1\times1$ convolutional heads on a collection of dense downstream tasks, ensuring no spatial context and pixel-wise feature and structure matching.

We emphasize that deformations strongly restrict the output, forbidding feature editing and preventing task leakage, or neglect of specific information correlating to the tasks used or not used during training. Indeed, when training  without the depth and the segmentation tasks, performance remains strong on these held-out tasks. See supplementary materials for more detail. See Supplementary material for visualizations of deformations through upscaling process.
\subsection{Cross-Channel Consistency}
\label{subsec:cross_channel_consistency}

\begin{figure*}
    \centering
    \includegraphics[trim={0 0 .5cm 7cm},clip,width=1\linewidth]{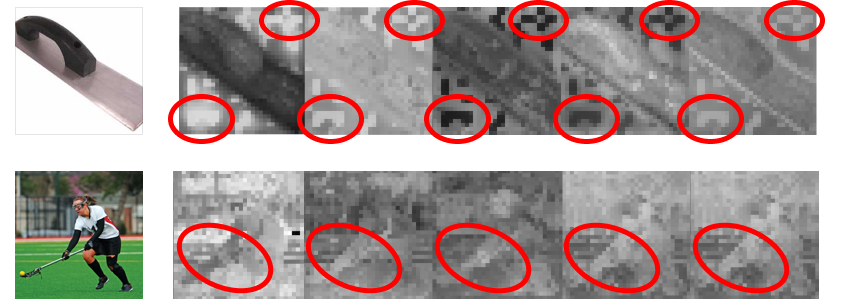}
    \includegraphics[trim={0 5.5cm 0 0cm},clip,width=.54\linewidth]{cross-channel_artifacts.png}
    \includegraphics[width=.45\linewidth]{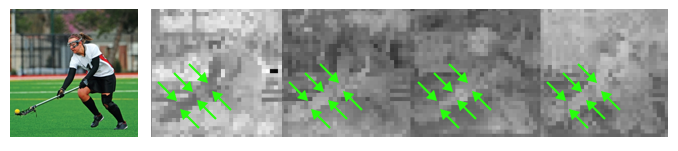}
    \caption{Visualizations of features values across different channels. We observe artifacts are consistent across channels (top row). In addition, low frequency regions of the image emphasize positional artifacts (bottom left), and the low resolution nature of the features typically induce thickening of object edges (bottom right).}
    \label{fig:simultaneous}
    \label{fig:artifacts}
\end{figure*}

Due to their patch-wise decomposition of the image and combined use of positional encodings, ViTs typically introduce artifacts into the produced  feature maps. We identify two main types of artifacts (Figure \ref{fig:artifacts}, bottom left).
First, in regions of near-constant color, positional artifacts are prominent and values may vary widely, without semantic difference. It has been observed \cite{darcet2024visiontransformersneedregisters} that this happens in great part from global semantics being locally encoded, but positional encodings are also a significant contributor, as they are the only varying input to the ViT in a constant patch.
Second is feature bleeding (Figure \ref{fig:artifacts}, bottom right): objects and boundaries are dilated due to discretization.

We make the crucial observation that ViTs introduce artifacts in a way which is consistent across channels (Figure~\ref{fig:simultaneous}, top). To substantiate this, we look at edge correlation between channels, and the original input image: we analyze 1{,}000 random ImageNet images. In each image, we isolate near-constant regions with an unsupervised local-variance filter, followed by erosion at the patch-grid scale so that measured gradients are not contaminated by neighboring edges; images without a sufficiently large flat region are discarded, leaving about $10\%$ of images. This selection is purely geometric and independent of the correlations being measured. On the remaining patches, we compute Sobel gradients of the backbone feature map and input image, and report the mean cross-channel Spearman correlation of gradient magnitudes, alongside two baselines: the same cross-channel correlation over the full token grid, and the channel-to-image correlation within the same masked region. In flat regions, cross-channel agreement is substantially higher than channel-to-image structure for both DINOv2 (0.29 vs.\ 0.11) and DINOv3 (0.30 vs.\ 0.12), and exceeds the full-image cross-channel baseline (0.18 and 0.22), showing that channel-aligned artifacts are concentrated where image gradients are weak rather than being a generic property of the grid. As a control, at image edges the channel-to-image correlation rises and the gap closes (0.17 vs.\ 0.15 for DINOv2; 0.19 vs.\ 0.23 for DINOv3), as expected when feature gradients follow real boundaries. Gradient directions are likewise shared across channels in flat regions (mean 
$\cos^2\approx0.73$ for both backbones). These findings indicate that one can learn a correction to all channels simultaneously, leading to very efficient resolution recovery.

\subsection{Architecture}
\label{sec:architecture}

\begin{figure*}
    \centering
    \includegraphics[width=.9\linewidth]{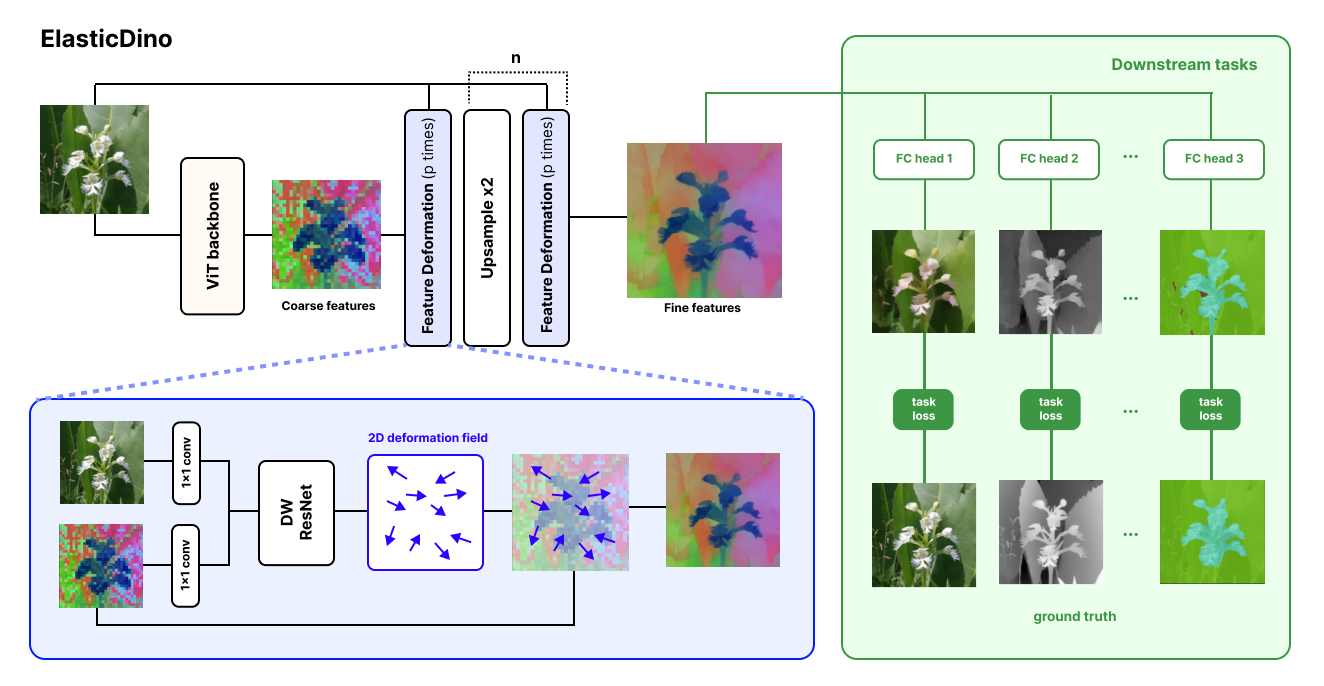}
    \caption{Method overview. The input image is passed to the (DINO) backbone for the low resolution, artifact-full features. These are successively passed to a sequence feature-fixing layers and then bilinearly upsampled by a factor 2. \textbf{Feature Deformation Layers} project and concatenate both the image and the semantic features, which are fed to a series of depthwise residual blocks, that end up generating a deformation field that warps the feature map. \textbf{Task heads} induce the sole losses of the system. The are simple MLPs that operate on the individual pixel level. Each head is compared to the ground truth using a task-specific loss.}
    \label{fig:architecture}
\end{figure*}

The architecture, depicted in Fig. \ref{fig:architecture} consists of alternating Feature Deformation layers, which correct artifacts and 2x bilinear upsamples. A first deformation layer is applied, then followed by an alternation of upsampling and deformations.
The deformation process is guided by the input image, allowing to introduce the image structure at that resolution to guide the deformation of the features.

Each upscaling stage is composed of a series of Feature Deformation Blocks, which take as inputs the "dirty" features at the current resolution and the image at the same resolution. These are projected to a common hidden dimension and concatenated; and then fed to a depthwise-separable \citep{chollet2017xceptiondeeplearningdepthwise} residual convolution network, which outputs a 2D deformation field. The field is applied to the input features to recover the denoised features.

In other terms, given features $f$, an image $I$ at a current identical resolution, a single deformation block consists of two projections $p_f$, $p_I$, a resnet encoder $R$ which outputs a 2D deformation field

$$\vec{F}=R(p_f(f)\oplus p_I(I))$$

Which is applied coordinate-wise:

$$\hat{f}_{u,v} = f_{\vec{F}_{u,v}}$$

The motivation for using depthwise separable convolutions is the desire to achieve an extremely lightweight architecture, given the fact that feature dimensions are already high, which at higher resolution leads to heavy memory and compute utilization. We observed no performance degradation as well as much faster training and inference.


\paragraph{Pixel-wise downstream tasks for semantic guidance}

Unfortunately, we have no ground truth maps in the upscaled resolution, and thus need to define an alternate training objective.

We make the observation that a well-upscaled and denoised feature map should be able, by design, to predict any dense task pixel-wise.

We hence introduce a multi-task learning approach, where we simultaneously train linear heads for downstream dense vision tasks (see Section~\ref{sec:losses}). Crucially, the 1x1 receptive field prevents the heads from combining feature values from different spatial positions, forcing the predictions to only use the semantics at that exact pixel. Should a larger receptive field be employed, the incentive to match the semantics at the current pixel would be removed, and structure would be learnt both in the upscaler and the downstream head.

It is important to note that this approach can only work because we are employing deformations and sampling from the features themselves. Indeed, if the upsampling process were to directly edit feature values, the semantics of the downstream tasks could easily just become "baked in" to the feature maps, without any incentive to jointly edit all channels simultaneously. We show the effect of such an approach in an ablation study- Fig. \ref{fig:ablations}.

For each downstream task (see below), we employ a simple linear head to predict the target. Furthermore, to encourage faster learning at every resolution, the same head is applied to every intermediate feature map, and the corresponding multi-resolution predictions are all used to compute the training objective.

\subsection{Losses and Tasks}
\label{sec:losses}

All the tasks we use predict a dense map on the full resolution. For every task except segmentation (for which we use the standard cross-entropy loss), the base loss is defined as $\mathcal{L}_{\textrm{base}}(x, \hat{x}) = ||x - \hat{x}||^{2} + \textrm{SSIM}(x, \hat{x})$, where $\textrm{SSIM}$ denotes the structural similarity loss \citep{wang2004image}.
For all tasks except segmentation, and in order to encourage reconstruction of high frequency details, we additionally penalize discrepancies between the channel-wise spatial gradients and laplacians, following \cite{bochkovskii2025depthprosharpmonocular}: $
\mathcal{L}_{\textrm{grad}}(x, \hat{x})= \mathcal{L}_{\textrm{base}}(\nabla x, \nabla \hat{x}), \, \mathcal{L}_{\textrm{lap}}(x, \hat{x}) = \mathcal{L}_{\textrm{base}}(\Delta x, \Delta \hat{x})$

To encourage faster reconstruction at every stage, we compute the loss for all intermediary feature maps, as well as at every resolution. Therefore, denoting $\hat{x}_{i, j}$ the output of the $i$-th stage for the ${j}$-th prediction head, $\hat{x}_j$ the ground truth for task $j$, $r_k = 32 \times 2^{k}$ where $k \in {[} {[} 0, 3 {]} {]}$ the resolutions used, and $\mathcal{L}_j$ the loss for task $j$ (see the task list below), we apply the following multi-scale loss:

$$\mathcal{L}_{\textrm{ms}}(x, \hat{x}) = \sum_{i, j, k} \mathcal{L}_{j}(\textrm{interpolate}(x_i, r_k), \textrm{interpolate}(\hat{x}y_{i, j}, r_k))$$

This allows gradients to flow at each stage, rather than through the entire upscaling network, greatly stabilizing and speeding up the learning. 


The downstream tasks we use represent a diverse sets of maps to be predicted from the image. We find each to help improve overall embedding quality (see Section~\ref{sec:experiments}), and believe more can be added. We use the following downstream tasks: RGB reconstruction, monocular depth, semantic segmentation, surface normals, intrinsic decomposition.
Training is performed on two interspersed datasets: Hypersim \citep{roberts2021hypersimphotorealisticsyntheticdataset} and Imagenet-22k \citep{deng2009imagenet}. See Supplementary material for details on losses and the training procedure.

\section{Experiments}\label{sec:experiments}

We perform quantitative and qualitative evaluation on a prototype designed to upscale DINOv2 features, and report the results below. We report semantic segmentation and depth estimation comparisons, and open vocabulary segmentation in the supplementary material. Following recent publications \cite{wimmer2025anyup}, we also perform some of the same experiments for DINOv3 \cite{simeoni2025dinov3} features. For implementation details, including results from our training head tasks, see supplementary material. 

\begin{figure*}[htbp!]
    \centering
    \includegraphics[width=\linewidth]{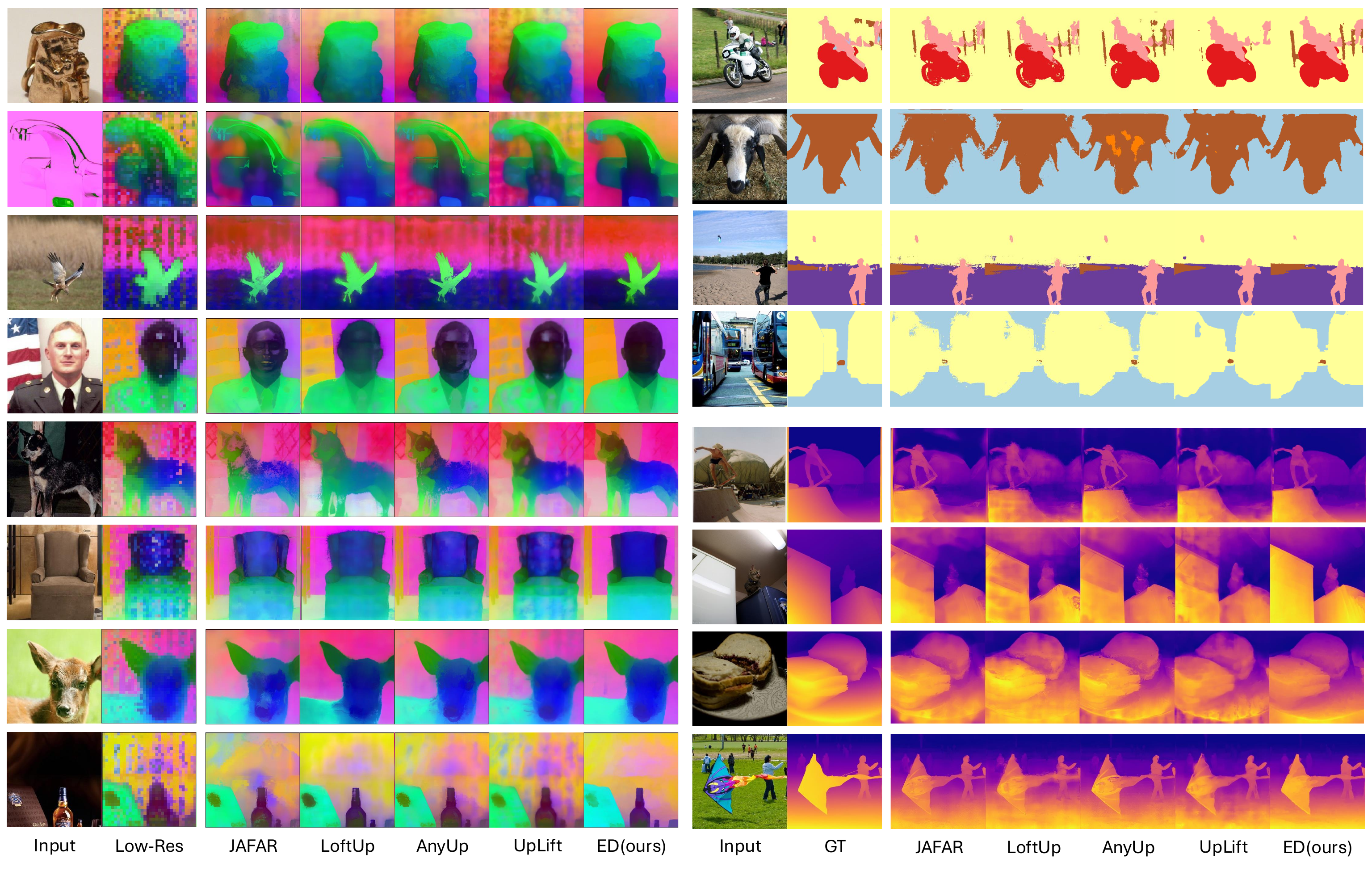}
    \caption{Qualitative evaluation. Left: The top three PCA components of the feature maps of input images are depicted, as generated originally by DINOv2 (Low-Res), previous approaches, and us. Top Right: Semantic segmentation maps of input images, as generated by humans (GT), three previous methods, and us are also depicted. Bottom right: depth maps for the same approaches and input setting are depicted as well. Throughout all results, \ourmethod depicts sharper edges that respect the input image resolution and mitigates semantic bleeding between regions, even when RGB colors are similar.}
    \label{fig:misc-results}
\end{figure*}

\textbf{Qualitative Results} are presented in Figure \ref{fig:misc-results}. Along side the input image, we visualize the top-3 PCA components of DINOv2 features, and resulting upscaling maps. As can be seen, artifacts are corrected and crisp semantic edges are formed, corresponding to the original image structure. \ourmethod alleviates artifacts in low variance regions of the image, and demonstrates semantic results that are decoupled from RGB variations in the image. For example, The similarly colored faucet is represented with distinctly different features compared to the background, while the flag is represented similarly throughout even though it consists of differently colored texture.  For a more comprehensive demonstration of qualitative results, please refer to the supplementary material. We also note that while positional encoding related artifacts are alleviated, the positional information itself persists, and is actually extracted from the features with more ease compared to competing upscaling methods. See supplementary material for details on this experiment. 

\subsection{Quantitative Comparisons}
\subsubsection{Task Specific (Semantic segmentation)}
We train simple convolution heads for semantic segmentation on top of \ourmethod and compare performance with two leading dedicated segmentation methods, Mask2Former \citep{cheng2021mask2former} and OneFormer \citep{jain2022oneformertransformerruleuniversal}. The data sets examined are MS-COCO \citep{lin2015microsoftcococommonobjects}, ADE20k \citep{zhou2017scene,zhou2018semantic} and Cityscapes \citep{cordts2016cityscapesdatasetsemanticurban} (Cscapes), all standard data sets in semantic segmentation benchmarks.

\ourmethod displays higher performance as measured by Mean-IoU (Intersection over Union) on all three validation datasets, with a very large margin on ADE20k and Cityscapes. Qualitatively (In Supplementary material) we observe not only better classification ability, but much finer segmentation boundaries. This implies obvious advantages to downstream applications where sensitivity is key, such as autonomous driving.

\subsubsection{Task Agnostic (Segmentation and Depth)} 
As our primary comparison, we further compare to task-agnostic methods that employ DINO feature upsampling (Table \ref{tab:main_table} for DINOv2, and Table \ref{tab:dino3_table} for available methods with DINOv3). These include training free Nearest neighbor and bilinear upsampling, and recent state-of-the-art methods \cite{fu2024featup, huang2025loftup,couairon2025jafar,wimmer2025anyup,walmer2026uplift}. We follow established evaluation protocols, as described in by JAFAR \cite{couairon2025jafar}, measuring mean Intersection over Union (mIoU) and pixel-level accuracy (Acc) for the semantic segmentation task over four datasets, and Root Mean Square Error (RMSE) and $\delta_1$ (measuring the  percentage of pixels where that are less than 25\% of the ground-truth) for the depth estimation task over one dataset. In the supplementary material, we also compare performance for zero-shot open vocabulary segmentation, and the CLIP \cite{radford2021learningtransferablevisualmodels} backbone. As can be seen, \ourmethod demonstrates leading performance consistently for all datasets except Cityscapes. We interpret this surprising discrepancy to the lower ability for our method to capture small objects, possibly due to the low-frequency bias of the convolutional architecture. To support this class size bias, we provide in the supplementary material a comparison of model performance by object instance size, demonstrating that while ElasticDino vastly outperforms other methods on larger classes, it struggles with smaller ones, indicating both better semantic preservation but also imperfect reconstruction of small objects.
Notably, these improved results are produced with less computational efforts and memory footprint, as reported in Table \ref{tab:performance}. 

\begin{table}
    \centering
    \caption{Forward pass performance measures of \ourmethod and competing feature upscaling works over DINOv2. Note that even though \ourmethod trains more parameters, it is faster and more memory efficient, due to the linear (as opposed to attention-based) nature of our operations}
    \resizebox{\columnwidth}{!}{
    \begin{tabular}{lccccc}
        \toprule
        & {ED (ours)} & {AnyUp} & {LoftUp} & {JAFAR} & {UPLift} \\
        \midrule
        Avg Time (ms) & 108.41 & 545.42 & 418.89 & 177.79 & 147.27 \\
        Peak mem (GB) & 0.71 & 1.94 & 8.31 & 7.33 & 9.22 \\
        \#Params (K) & {27,095} & {878} & {4,303} & {628} & {789} \\
        \bottomrule
    \end{tabular}
    }
    \label{tab:performance}
\end{table}

\begin{figure}
     \centering
\captionof{table}{Validation Mean IoU for semantic segmentation benchmarks. Note that results differ from the comparison benchmarks, as we use a convolutional head (rather than linear) and train for longer than specified in the JAFAR benchmark.}

\begin{tabular}[b]{lccc}

\toprule
\textbf{Model} & \textbf{Cscapes} & \textbf{COCO} & \textbf{ADE20k} \\

\midrule
Oneformer-sw-L &  \underline{0.6412} & \underline{0.6264} & \underline{0.5109} \\
Mask2Former-sw-L & 0.5986 & 0.6239 & 0.4928 \\
ElasticDino-L & \textbf{0.6875} & \textbf{0.6270} & \textbf{0.5299} \\
\bottomrule
\end{tabular}    
    \label{table:ablations}
\end{figure}

\begin{table*}[t]
\centering
\footnotesize
\setlength{\tabcolsep}{4pt}

\caption{Comparison of various upsampling methods. All benchmarks are run using the method and code used for the JAFAR paper. Bold = best, underline = second best.}
\label{tab:main_table}

\begin{adjustbox}{width=\textwidth}
\begin{tabular}{l c c c c c c c c c c}
\toprule

& \multicolumn{8}{c}{\textbf{Semantic Segmentation}}
& \multicolumn{2}{c}{\begin{tabular}[c]{@{}c@{}}\textbf{Depth Est.}\end{tabular}} \\

\cmidrule(lr){2-9}
\cmidrule(lr){10-11}

& \multicolumn{2}{c}{COCO}
& \multicolumn{2}{c}{VOC}
& \multicolumn{2}{c}{ADE20K}
& \multicolumn{2}{c}{Cityscapes}
& \multicolumn{2}{c}{COCO} \\

\cmidrule(lr){2-9}
\cmidrule(lr){10-11}

Method
& {mIoU$\uparrow$} & {Acc$\uparrow$}
& {mIoU$\uparrow$} & {Acc$\uparrow$}
& {mIoU$\uparrow$} & {Acc$\uparrow$}
& {mIoU$\uparrow$} & {Acc$\uparrow$}
& {$\delta_1\uparrow$} & {RMSE$\downarrow$} \\

\cmidrule(lr){1-1}
\cmidrule(lr){2-3}
\cmidrule(lr){4-5}
\cmidrule(lr){6-7}
\cmidrule(lr){8-9}
\cmidrule(lr){10-11}



Nearest
& 56.17 & 76.97
& 76.41 & 93.80
& 37.27 & 71.91
& 54.05 & 90.36
& 58.08 & 0.70 \\

Bilinear
& 59.03 & 79.07
& 80.70 & 95.17
& 39.23 & 73.69
& 59.37 & 92.47
& 59.92 & 0.66 \\









FeatUp \cite{fu2024featup}
& 60.10 & 79.95
& 81.08 & 95.32
& 38.82 & 73.74
& 56.06 & 91.86
& \underline{61.69} & \underline{0.64} \\

LoftUp \cite{huang2025loftup}
& \underline{61.62} & \underline{81.11}
& 84.18 & 96.22
& \underline{41.51} & \underline{75.41}
& \underline{62.13} & \underline{93.53}
& 58.01 & 0.68 \\

JAFAR \cite{couairon2025jafar}
& 60.32 & 80.30
& 84.19 & 96.18
& 40.16 & 74.39
& 61.58 & 93.45
& 60.05 & 0.65 \\

AnyUp \cite{wimmer2025anyup}
& 60.68 & 80.59
& 83.81 & 96.13
& 40.39 & 74.76
& 60.99 & 93.37
& 60.65 & \underline{0.64} \\

UPLift \cite{walmer2026uplift}
& 61.10 & 80.85
& \underline{84.82} & \underline{96.41}
& 41.29 & 75.05
& \textbf{65.02} & \textbf{94.34}
& 60.55 & \underline{0.64} \\

\cmidrule(lr){1-1}
\cmidrule(lr){2-3}
\cmidrule(lr){4-5}
\cmidrule(lr){6-7}
\cmidrule(lr){8-9}
\cmidrule(lr){10-11}

ED (ours)
& \textbf{62.72} & \textbf{81.74}
& \textbf{86.00} & \textbf{96.78}
& \textbf{42.10} & \textbf{76.00}
& 61.33 & 93.44
& \textbf{64.14} & \textbf{0.57} \\

\bottomrule
\end{tabular}
\end{adjustbox}

\vspace{1em}

\end{table*}

\begin{table*}
\centering
\footnotesize
\setlength{\tabcolsep}{4pt}

\caption{Comparison of models over DINOv3-small backbone. Benchmarks are run using the method and code used for the JAFAR \cite{couairon2025jafar} paper. Bold = best, underline = second best.}
\label{tab:dino3_table}
\begin{adjustbox}{width=\textwidth}
    \small 
    \begin{tabular}{l c c c c c c c c c c}
        \toprule
        \multirow{3}{*}{\textbf{DinoV3}} & \multicolumn{8}{c}{\textbf{Semantic Segmentation}} & \multicolumn{2}{c}{\textbf{Depth Est.}} \\
        \cmidrule(lr){2-9} \cmidrule(lr){10-11}
        & \multicolumn{2}{c}{COCO} & \multicolumn{2}{c}{VOC} & \multicolumn{2}{c}{ADE20K} & \multicolumn{2}{c}{Cityscapes} & \multicolumn{2}{c}{COCO} \\
        \cmidrule(lr){2-3} \cmidrule(lr){4-5} \cmidrule(lr){6-7} \cmidrule(lr){8-9} \cmidrule(lr){10-11}
        & mIoU($\uparrow$) & Acc($\uparrow$) & mIoU($\uparrow$) & Acc($\uparrow$) & mIoU($\uparrow$) & Acc($\uparrow$) & mIoU($\uparrow$) & Acc($\uparrow$) & $\delta_1$($\uparrow$) & RMSE($\downarrow$) \\
        \midrule
        JAFAR \cite{couairon2025jafar} & 62.2 & 81.38 & 82.9 & 95.96 &
        42.31 & 75.7 & 57.25 & 92.3 & 58.2 & 0.68 \\
        AnyUp \cite{wimmer2025anyup} & 62.70 & 81.69 & 83.88 & 96.21 & 42.69 & 76.02 & 58.32 & 92.81 & \underline{59.43} & \underline{0.66} \\
        UPLift \cite{walmer2026uplift} &
        \underline{63.52} & \underline{82.14} & \underline{84.86} & \underline{96.58} &
        \underline{44.12} & \underline{76.62} & \underline{62.31} & \underline{93.82} & 59.29 & 0.66 \\
        ED (ours) & \textbf{64.04} & \textbf{82.43} & \textbf{86.26} & \textbf{96.89} & \textbf{44.57} & \textbf{77.03} & \textbf{62.91} & \textbf{94.03} & \textbf{61.10} & \textbf{0.62} \\
        \bottomrule
    \end{tabular}
\end{adjustbox}
\end{table*}

\subsection{Ablation studies}

We perform an ablation study on the architecture and design choices. In order to ensure a fair benchmark, we train ElasticDino and its ablations on the Hypersim dataset and omit the depth task, reserving it for evaluation. The training tasks are RGB, illumination, reflectance, residuals and normals \cite{roberts2021hypersimphotorealisticsyntheticdataset}. 

First, we ablate the deformations, employing convolutions only. Second, we ablate tasks in a leave-one-out fashion. Finally, we ablate by progressively omitting tasks. 

\begin{figure*}[t]
\centering

\begin{minipage}[t]{0.35\textwidth}
    \centering
    \vspace{0pt}
    \includegraphics[width=\linewidth]{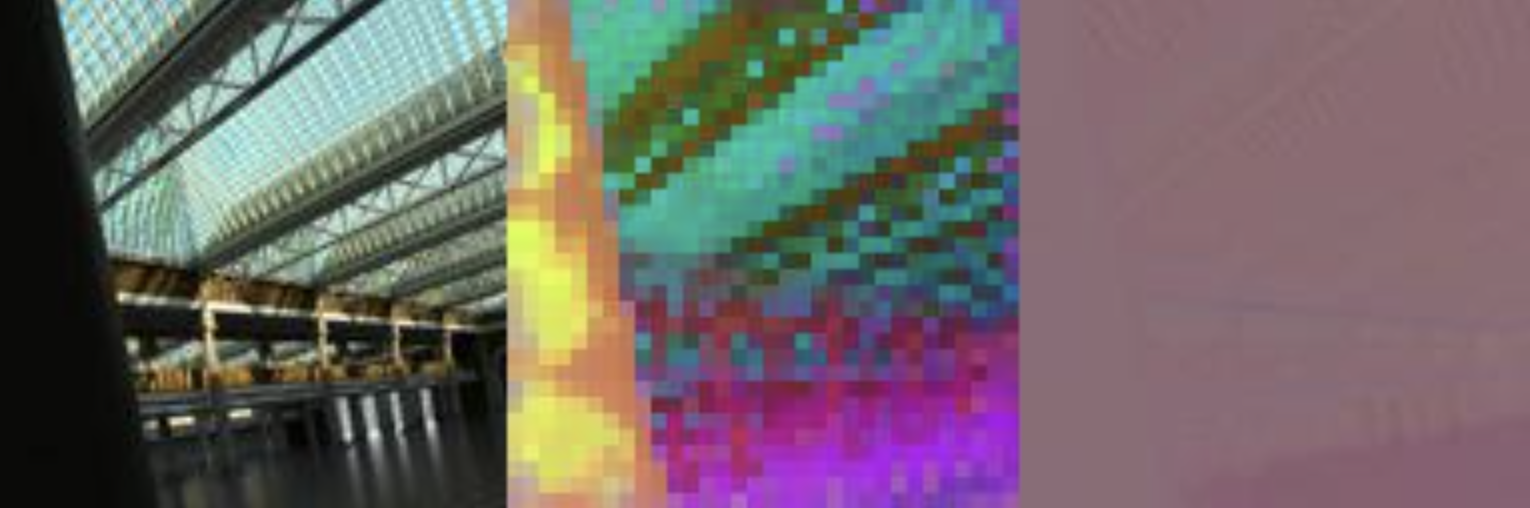}

    \captionof{figure}{
        Ablating deformations weakens the preservation of original features by allowing the encoding of the task specific values in the output, instead of only blending them.
        From left to right: input image, low-resolution DINOv2 features,
        the result without deformations.
    }
    \label{fig:ablations}
\end{minipage}
\hfill
\begin{minipage}[t]{0.63\textwidth}
    \centering
    \vspace{0pt}

    \captionof{table}{Depth-estimation ablation test loss}
    \label{tab:depth-ablations}
    \vspace{-1.5mm}

    \scriptsize
    \setlength{\tabcolsep}{2.0pt}
    \renewcommand{\arraystretch}{0.95}

    \begin{adjustbox}{max width=\linewidth}
    \begin{tabular}{@{}l*{9}{c}@{}}
        \toprule
        Ablation
        & $\delta_1(\%)\uparrow$
        & $\Delta\delta_1$
        & $\delta_3(\%)\uparrow$
        & AbsRel$\downarrow$
        & SqRel$\downarrow$
        & RMSE$\downarrow$
        & $\Delta$RMSE
        & $\log_{10}\downarrow$
        & SILog$\downarrow$ \\
        \midrule

        w/o DEFormations
        & 26.80
        & $-31.55$
        & 70.11
        & 1.024
        & 1.19
        & 1.32
        & $+0.718$
        & 0.238
        & 66.30 \\

        w/o ILLumination
        & 57.21
        & $-1.14$
        & 92.83
        & 0.428
        & 0.278
        & 0.620
        & $+0.018$
        & 0.117
        & 33.06 \\

        w/o REFlectance
        & \textbf{58.43}
        & $+0.08$
        & 93.10
        & 0.419
        & 0.268
        & 0.603
        & $+0.001$
        & \underline{0.115}
        & \textbf{32.45} \\

        w/o RESiduals
        & 57.74
        & $-0.61$
        & 93.02
        & 0.423
        & 0.273
        & 0.611
        & $+0.009$
        & 0.116
        & 32.73 \\

        w/o Normals
        & 57.81
        & $-0.54$
        & 92.95
        & 0.424
        & 0.274
        & 0.611
        & $+0.009$
        & 0.116
        & 32.89 \\

        w/o RGB
        & 58.22
        & $-0.13$
        & \underline{93.13}
        & \underline{0.418}
        & \underline{0.266}
        & \underline{0.603}
        & $+0.001$
        & \underline{0.115}
        & 32.47 \\

        \midrule

        w/o RGB,ILL
        & 57.20
        & $-1.15$
        & 92.81
        & 0.425
        & 0.273
        & 0.617
        & $+0.015$
        & 0.117
        & 33.24 \\

        w/o RGB,ILL,REF
        & 57.17
        & $-1.18$
        & 92.73
        & 0.432
        & 0.285
        & 0.622
        & $+0.020$
        & 0.118
        & 33.54 \\

        w/o RGB,ILL,REF,RES
        & 56.29
        & $-2.06$
        & 92.50
        & 0.437
        & 0.292
        & 0.634
        & $+0.032$
        & 0.120
        & 33.99 \\

        \midrule

        full model
        & \underline{58.35}
        & --
        & \textbf{93.17}
        & \textbf{0.417}
        & \textbf{0.265}
        & \textbf{0.602}
        & --
        & \textbf{0.115}
        & \underline{32.46} \\

        \bottomrule
    \end{tabular}
    \end{adjustbox}

    \vspace{0.6mm}

    Bold denotes the best result and Underline the second-best result. Rankings are computed from the unrounded values; consequently, identically displayed rounded values may have different ranking markers.
\end{minipage}

\end{figure*}

For each trained upscaler, we train a UNet head to predict the unseen depth, reserving a fraction as a test set (which is also held out from upscaler training), with SiLogLoss as the depth metric. Quantitative results are summarized in table \ref{tab:depth-ablations}, and some qualitative results are shown in figure \ref{fig:ablations}. 

The ablation of deformations results in almost imperceptible differences in feature values: because of the lack of incentive for the network to jointly modify features, task semantics are leaked and baked into the feature values. Deformation-based upsampling preserves feature semantics, allowing only local feature resampling and alleviating catastrophic forgetting. When deformations are ablated, the upsampler can directly alter feature values and may encode only the supervised task signal, e.g. depth, rather than preserving semantic structure. However, because some tasks exhibit correlation with depth prediction (e.g. surface normals), and because residual connections let the features flow through, some ability to transfer is conserved. It can be seen that the model gets to good performance with low amount of tasks but still improves with additional tasks. When ablating each task individually, we see that every task leads to some amount of improvement in performance, showing that adding more tasks improves the performance.


\setlength{\columnsep}{2pt}%
\begin{wrapfigure}[5]{r}{1.5in}
  \centering
    \vspace{-1.1em}

    \includegraphics[width=.45\columnwidth]{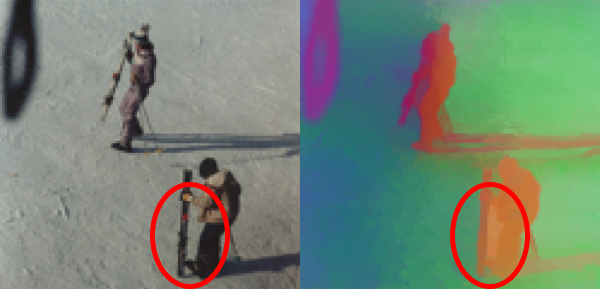}
    \label{fig:blocking}
\end{wrapfigure}

\section{Discussion}\label{sec:analysis}

\subsection{Limitations}

While demonstrating promising performance, \ourmethod  struggles with distinguishing small holes in the source image that are smaller than patch size (see inset figure). Our intuition is that the convolutional layers still display low frequency biases especially at large receptive fields, even though leakage is mitigated due to the deformation. Further experimenting with alternative architectures with lower frequency bias may be rewarding.

\subsection{Future directions}\label{sec:future}

Beyond solving the above-mentioned issue, a clear future direction is the extension of ElasticDino to other modalities, such as video or 3D. Video could be handled through video-specific feature generators such as VJEPA-2 \cite{assran2025vjepa2selfsupervisedvideo}. 3D could be handled through multi-view projection of features from renderings, as done in TRELLIS \cite{xiang2025structured3dlatentsscalable}.

\subsection{Conclusion}\label{sec:conclusion}

We have introduced \ourmethod, a solution for producing high resolution semantic features on top of vision transformers. We demonstrated \ourmethod achieves superior performance compared to existing methods, with lower memory usage and faster runtimes.

ElasticDino, like the vision transformers it enhances, does not produce content that can be directly misused. As a foundational model, any broader impact, positive or negative, is likely to stem from improved performance in downstream tasks or the emergence of new applications.

We refer to the supplementary material for additional implementation details and qualitative results.
{
    \small
    \bibliography{main}

\begin{thebibliography}{40}
\providecommand{\natexlab}[1]{#1}
\providecommand{\url}[1]{\texttt{#1}}
\expandafter\ifx\csname urlstyle\endcsname\relax
  \providecommand{\doi}[1]{doi: #1}\else
  \providecommand{\doi}{doi: \begingroup \urlstyle{rm}\Url}\fi

\bibitem[Assran et~al.(2025)Assran, Bardes, Fan, Garrido, Howes, Mojtaba,
  Komeili, Muckley, Rizvi, Roberts, Sinha, Zholus, Arnaud, Gejji, Martin,
  Hogan, Dugas, Bojanowski, Khalidov, Labatut, Massa, Szafraniec, Krishnakumar,
  Li, Ma, Chandar, Meier, LeCun, Rabbat, and
  Ballas]{assran2025vjepa2selfsupervisedvideo}
Mido Assran, Adrien Bardes, David Fan, Quentin Garrido, Russell Howes, Mojtaba,
  Komeili, Matthew Muckley, Ammar Rizvi, Claire Roberts, Koustuv Sinha, Artem
  Zholus, Sergio Arnaud, Abha Gejji, Ada Martin, Francois~Robert Hogan, Daniel
  Dugas, Piotr Bojanowski, Vasil Khalidov, Patrick Labatut, Francisco Massa,
  Marc Szafraniec, Kapil Krishnakumar, Yong Li, Xiaodong Ma, Sarath Chandar,
  Franziska Meier, Yann LeCun, Michael Rabbat, and Nicolas Ballas.
\newblock V-jepa 2: Self-supervised video models enable understanding,
  prediction and planning, 2025.

\bibitem[Bochkovskii et~al.(2025)Bochkovskii, Delaunoy, Germain, Santos, Zhou,
  Richter, and Koltun]{bochkovskii2025depthprosharpmonocular}
Aleksei Bochkovskii, Amaël Delaunoy, Hugo Germain, Marcel Santos, Yichao Zhou,
  Stephan~R. Richter, and Vladlen Koltun.
\newblock Depth pro: Sharp monocular metric depth in less than a second, 2025.

\bibitem[Caron et~al.(2021)Caron, Touvron, Misra, J{\'e}gou, Mairal,
  Bojanowski, and Joulin]{caron2021emerging}
Mathilde Caron, Hugo Touvron, Ishan Misra, Herv{\'e} J{\'e}gou, Julien Mairal,
  Piotr Bojanowski, and Armand Joulin.
\newblock Emerging properties in self-supervised vision transformers.
\newblock In \emph{Proceedings of the IEEE/CVF international conference on
  computer vision}, pages 9650--9660, 2021.

\bibitem[Chen et~al.(2023)Chen, Duan, Wang, He, Lu, Dai, and
  Qiao]{chen2023visiontransformeradapterdense}
Zhe Chen, Yuchen Duan, Wenhai Wang, Junjun He, Tong Lu, Jifeng Dai, and Yu
  Qiao.
\newblock Vision transformer adapter for dense predictions, 2023.

\bibitem[Cheng et~al.(2021)Cheng, Misra, Schwing, Kirillov, and
  Girdhar]{cheng2021mask2former}
Bowen Cheng, Ishan Misra, Alexander~G. Schwing, Alexander Kirillov, and Rohit
  Girdhar.
\newblock Masked-attention mask transformer for universal image segmentation.
\newblock \emph{arXiv}, 2021.

\bibitem[Chollet(2017)]{chollet2017xceptiondeeplearningdepthwise}
François Chollet.
\newblock Xception: Deep learning with depthwise separable convolutions, 2017.

\bibitem[Cordts et~al.(2016)Cordts, Omran, Ramos, Rehfeld, Enzweiler, Benenson,
  Franke, Roth, and Schiele]{cordts2016cityscapesdatasetsemanticurban}
Marius Cordts, Mohamed Omran, Sebastian Ramos, Timo Rehfeld, Markus Enzweiler,
  Rodrigo Benenson, Uwe Franke, Stefan Roth, and Bernt Schiele.
\newblock The cityscapes dataset for semantic urban scene understanding, 2016.

\bibitem[Couairon et~al.(2025)Couairon, Chambon, Serrano, Haugeard, Cord, and
  Thome]{couairon2025jafar}
Paul Couairon, Loick Chambon, Louis Serrano, Jean-Emmanuel Haugeard, Matthieu
  Cord, and Nicolas Thome.
\newblock Jafar: Jack up any feature at any resolution, 2025.

\bibitem[Darcet et~al.(2024)Darcet, Oquab, Mairal, and
  Bojanowski]{darcet2024visiontransformersneedregisters}
Timothée Darcet, Maxime Oquab, Julien Mairal, and Piotr Bojanowski.
\newblock Vision transformers need registers, 2024.

\bibitem[Deng et~al.(2009)Deng, Dong, Socher, Li, Li, and
  Fei-Fei]{deng2009imagenet}
Jia Deng, Wei Dong, Richard Socher, Li-Jia Li, Kai Li, and Li Fei-Fei.
\newblock Imagenet: A large-scale hierarchical image database.
\newblock In \emph{2009 IEEE conference on computer vision and pattern
  recognition}, pages 248--255. Ieee, 2009.

\bibitem[Fu et~al.(2024)Fu, Hamilton, Brandt, Feldmann, Zhang, and
  Freeman]{fu2024featup}
Stephanie Fu, Mark Hamilton, Laura~E. Brandt, Axel Feldmann, Zhoutong Zhang,
  and William~T. Freeman.
\newblock Featup: A model-agnostic framework for features at any resolution.
\newblock In \emph{The Twelfth International Conference on Learning
  Representations}, 2024.

\bibitem[He et~al.(2015)He, Zhang, Ren, and
  Sun]{he2015deepresiduallearningimage}
Kaiming He, Xiangyu Zhang, Shaoqing Ren, and Jian Sun.
\newblock Deep residual learning for image recognition, 2015.

\bibitem[Huang et~al.(2025)Huang, Chen, Havrylov, Geiger, and
  Zhang]{huang2025loftup}
Haiwen Huang, Anpei Chen, Volodymyr Havrylov, Andreas Geiger, and Dan Zhang.
\newblock Loftup: Learning a coordinate-based feature upsampler for vision
  foundation models.
\newblock \emph{arXiv preprint arXiv:2504.14032}, 2025.

\bibitem[Jain et~al.(2022)Jain, Li, Chiu, Hassani, Orlov, and
  Shi]{jain2022oneformertransformerruleuniversal}
Jitesh Jain, Jiachen Li, MangTik Chiu, Ali Hassani, Nikita Orlov, and Humphrey
  Shi.
\newblock Oneformer: One transformer to rule universal image segmentation,
  2022.

\bibitem[Lin et~al.(2015)Lin, Maire, Belongie, Bourdev, Girshick, Hays, Perona,
  Ramanan, Zitnick, and Dollár]{lin2015microsoftcococommonobjects}
Tsung-Yi Lin, Michael Maire, Serge Belongie, Lubomir Bourdev, Ross Girshick,
  James Hays, Pietro Perona, Deva Ramanan, C.~Lawrence Zitnick, and Piotr
  Dollár.
\newblock Microsoft coco: Common objects in context, 2015.

\bibitem[Liu et~al.(2024)Liu, Zeng, Ren, Li, Zhang, Yang, Jiang, Li, Yang, Su,
  Zhu, and Zhang]{liu2024groundingdinomarryingdino}
Shilong Liu, Zhaoyang Zeng, Tianhe Ren, Feng Li, Hao Zhang, Jie Yang, Qing
  Jiang, Chunyuan Li, Jianwei Yang, Hang Su, Jun Zhu, and Lei Zhang.
\newblock Grounding dino: Marrying dino with grounded pre-training for open-set
  object detection, 2024.

\bibitem[Liu et~al.(2023)Liu, Lu, Fu, and Cao]{liu2023learning}
Wenze Liu, Hao Lu, Hongtao Fu, and Zhiguo Cao.
\newblock Learning to upsample by learning to sample.
\newblock In \emph{Proceedings of the IEEE/CVF international conference on
  computer vision}, pages 6027--6037, 2023.

\bibitem[Lu et~al.(2022)Lu, Liu, Ye, Fu, Liu, and Cao]{lu2022sapa}
Hao Lu, Wenze Liu, Zixuan Ye, Hongtao Fu, Yuliang Liu, and Zhiguo Cao.
\newblock Sapa: Similarity-aware point affiliation for feature upsampling.
\newblock \emph{Advances in Neural Information Processing Systems},
  35:\penalty0 20889--20901, 2022.

\bibitem[Oquab et~al.(2023)Oquab, Darcet, Moutakanni, Vo, Szafraniec, Khalidov,
  Fernandez, Haziza, Massa, El-Nouby, et~al.]{oquab2023dinov2}
Maxime Oquab, Timoth{\'e}e Darcet, Th{\'e}o Moutakanni, Huy Vo, Marc
  Szafraniec, Vasil Khalidov, Pierre Fernandez, Daniel Haziza, Francisco Massa,
  Alaaeldin El-Nouby, et~al.
\newblock Dinov2: Learning robust visual features without supervision.
\newblock \emph{arXiv preprint arXiv:2304.07193}, 2023.

\bibitem[Peebles and Xie(2023)]{peebles2023scalablediffusionmodelstransformers}
William Peebles and Saining Xie.
\newblock Scalable diffusion models with transformers, 2023.

\bibitem[Radford et~al.(2021)Radford, Kim, Hallacy, Ramesh, Goh, Agarwal,
  Sastry, Askell, Mishkin, Clark, Krueger, and
  Sutskever]{radford2021learningtransferablevisualmodels}
Alec Radford, Jong~Wook Kim, Chris Hallacy, Aditya Ramesh, Gabriel Goh,
  Sandhini Agarwal, Girish Sastry, Amanda Askell, Pamela Mishkin, Jack Clark,
  Gretchen Krueger, and Ilya Sutskever.
\newblock Learning transferable visual models from natural language
  supervision, 2021.

\bibitem[Ranftl et~al.(2020)Ranftl, Lasinger, Hafner, Schindler, and
  Koltun]{ranftl2020robustmonoculardepthestimation}
René Ranftl, Katrin Lasinger, David Hafner, Konrad Schindler, and Vladlen
  Koltun.
\newblock Towards robust monocular depth estimation: Mixing datasets for
  zero-shot cross-dataset transfer, 2020.

\bibitem[Ranftl et~al.(2021)Ranftl, Bochkovskiy, and
  Koltun]{ranftl2021visiontransformersdenseprediction}
René Ranftl, Alexey Bochkovskiy, and Vladlen Koltun.
\newblock Vision transformers for dense prediction, 2021.

\bibitem[Roberts et~al.(2021)Roberts, Ramapuram, Ranjan, Kumar, Bautista,
  Paczan, Webb, and
  Susskind]{roberts2021hypersimphotorealisticsyntheticdataset}
Mike Roberts, Jason Ramapuram, Anurag Ranjan, Atulit Kumar, Miguel~Angel
  Bautista, Nathan Paczan, Russ Webb, and Joshua~M. Susskind.
\newblock Hypersim: A photorealistic synthetic dataset for holistic indoor
  scene understanding, 2021.

\bibitem[Sim{\'e}oni et~al.(2025)Sim{\'e}oni, Vo, Seitzer, Baldassarre, Oquab,
  Jose, Khalidov, Szafraniec, Yi, Ramamonjisoa, Massa, Haziza, Wehrstedt, Wang,
  Darcet, Moutakanni, Sentana, Roberts, Vedaldi, Tolan, Brandt, Couprie,
  Mairal, J{\'e}gou, Labatut, and Bojanowski]{simeoni2025dinov3}
Oriane Sim{\'e}oni, Huy~V. Vo, Maximilian Seitzer, Federico Baldassarre, Maxime
  Oquab, Cijo Jose, Vasil Khalidov, Marc Szafraniec, Seungeun Yi, Micha{\"e}l
  Ramamonjisoa, Francisco Massa, Daniel Haziza, Luca Wehrstedt, Jianyuan Wang,
  Timoth{\'e}e Darcet, Th{\'e}o Moutakanni, Leonel Sentana, Claire Roberts,
  Andrea Vedaldi, Jamie Tolan, John Brandt, Camille Couprie, Julien Mairal,
  Herv{\'e} J{\'e}gou, Patrick Labatut, and Piotr Bojanowski.
\newblock {DINOv3}, 2025.

\bibitem[Suri et~al.(2024)Suri, Walmer, Gupta, and
  Shrivastava]{suri2024liftsurprisinglysimplelightweight}
Saksham Suri, Matthew Walmer, Kamal Gupta, and Abhinav Shrivastava.
\newblock Lift: A surprisingly simple lightweight feature transform for dense
  vit descriptors, 2024.

\bibitem[Tumanyan et~al.(2024)Tumanyan, Singer, Bagon, and
  Dekel]{tumanyan2024dinotrackertamingdinoselfsupervised}
Narek Tumanyan, Assaf Singer, Shai Bagon, and Tali Dekel.
\newblock Dino-tracker: Taming dino for self-supervised point tracking in a
  single video, 2024.

\bibitem[Walmer et~al.(2026)Walmer, Suri, Aggarwal, and
  Shrivastava]{walmer2026uplift}
Matthew Walmer, Saksham Suri, Anirud Aggarwal, and Abhinav Shrivastava.
\newblock Uplift: Efficient pixel-dense feature upsampling with local
  attenders.
\newblock \emph{arXiv preprint arXiv:2601.17950}, 2026.

\bibitem[Wang et~al.(2004)Wang, Bovik, Sheikh, and Simoncelli]{wang2004image}
Zhou Wang, Alan~Conrad Bovik, Hamid~Rahim Sheikh, and Eero~P. Simoncelli.
\newblock Image quality assessment: From error visibility to structural
  similarity.
\newblock \emph{IEEE Transactions on Image Processing}, 13\penalty0
  (4):\penalty0 600--612, 2004.

\bibitem[Wang et~al.(2022)Wang, Yu, Yu, Dai, Tsvetkov, and
  Cao]{wang2022simvlmsimplevisuallanguage}
Zirui Wang, Jiahui Yu, Adams~Wei Yu, Zihang Dai, Yulia Tsvetkov, and Yuan Cao.
\newblock Simvlm: Simple visual language model pretraining with weak
  supervision, 2022.

\bibitem[Wimmer et~al.(2025)Wimmer, Truong, Rakotosaona, Oechsle, Tombari,
  Schiele, and Lenssen]{wimmer2025anyup}
Thomas Wimmer, Prune Truong, Marie-Julie Rakotosaona, Michael Oechsle, Federico
  Tombari, Bernt Schiele, and Jan~Eric Lenssen.
\newblock Anyup: Universal feature upsampling.
\newblock \emph{arXiv preprint arXiv:2510.12764}, 2025.

\bibitem[Xiang et~al.(2025)Xiang, Lv, Xu, Deng, Wang, Zhang, Chen, Tong, and
  Yang]{xiang2025structured3dlatentsscalable}
Jianfeng Xiang, Zelong Lv, Sicheng Xu, Yu Deng, Ruicheng Wang, Bowen Zhang,
  Dong Chen, Xin Tong, and Jiaolong Yang.
\newblock Structured 3d latents for scalable and versatile 3d generation, 2025.

\bibitem[Yang et~al.(2024{\natexlab{a}})Yang, Luo, Li, Deng, Guibas, Krishnan,
  Weinberger, Tian, and Wang]{yang2024denoisingvisiontransformers}
Jiawei Yang, Katie~Z Luo, Jiefeng Li, Congyue Deng, Leonidas Guibas, Dilip
  Krishnan, Kilian~Q Weinberger, Yonglong Tian, and Yue Wang.
\newblock Denoising vision transformers, 2024{\natexlab{a}}.

\bibitem[Yang et~al.(2024{\natexlab{b}})Yang, Kang, Huang, Xu, Feng, and
  Zhao]{yang2024depthanythingunleashingpower}
Lihe Yang, Bingyi Kang, Zilong Huang, Xiaogang Xu, Jiashi Feng, and Hengshuang
  Zhao.
\newblock Depth anything: Unleashing the power of large-scale unlabeled data,
  2024{\natexlab{b}}.

\bibitem[Yang et~al.(2024{\natexlab{c}})Yang, Kang, Huang, Zhao, Xu, Feng, and
  Zhao]{yang2024depthv2}
Lihe Yang, Bingyi Kang, Zilong Huang, Zhen Zhao, Xiaogang Xu, Jiashi Feng, and
  Hengshuang Zhao.
\newblock Depth anything v2, 2024{\natexlab{c}}.

\bibitem[Yun et~al.(2022)Yun, Lee, Kim, and
  Shin]{yun2022patchlevelrepresentationlearningselfsupervised}
Sukmin Yun, Hankook Lee, Jaehyung Kim, and Jinwoo Shin.
\newblock Patch-level representation learning for self-supervised vision
  transformers, 2022.

\bibitem[Zhai et~al.(2022)Zhai, Kolesnikov, Houlsby, and
  Beyer]{zhai2022scalingvisiontransformers}
Xiaohua Zhai, Alexander Kolesnikov, Neil Houlsby, and Lucas Beyer.
\newblock Scaling vision transformers, 2022.

\bibitem[Zhou et~al.(2017)Zhou, Zhao, Puig, Fidler, Barriuso, and
  Torralba]{zhou2017scene}
Bolei Zhou, Hang Zhao, Xavier Puig, Sanja Fidler, Adela Barriuso, and Antonio
  Torralba.
\newblock Scene parsing through ade20k dataset.
\newblock In \emph{Proceedings of the IEEE Conference on Computer Vision and
  Pattern Recognition (CVPR)}, 2017.

\bibitem[Zhou et~al.(2018)Zhou, Zhao, Puig, Xiao, Fidler, Barriuso, and
  Torralba]{zhou2018semantic}
Bolei Zhou, Hang Zhao, Xavier Puig, Tete Xiao, Sanja Fidler, Adela Barriuso,
  and Antonio Torralba.
\newblock Semantic understanding of scenes through ade20k dataset.
\newblock \emph{International Journal of Computer Vision (IJCV)}, 2018.

\bibitem[Zhou et~al.(2025)Zhou, Wang, Zheng, and Meng]{zhou2025refreshed}
Minghao Zhou, Hong Wang, Yefeng Zheng, and Deyu Meng.
\newblock A refreshed similarity-based upsampler for direct high-ratio feature
  upsampling.
\newblock \emph{IEEE Transactions on Neural Networks and Learning Systems},
  2025.

\end{thebibliography}


\begin{thebibliography}{18}
\providecommand{\natexlab}[1]{#1}
\providecommand{\url}[1]{\texttt{#1}}
\expandafter\ifx\csname urlstyle\endcsname\relax
  \providecommand{\doi}[1]{doi: #1}\else
  \providecommand{\doi}{doi: \begingroup \urlstyle{rm}\Url}\fi

\bibitem[Baker et~al.(2011)Baker, Scharstein, Lewis, Roth, Black, and
  Szeliski]{supp-baker2011database}
Simon Baker, Daniel Scharstein, James~P Lewis, Stefan Roth, Michael~J Black,
  and Richard Szeliski.
\newblock A database and evaluation methodology for optical flow.
\newblock \emph{International journal of computer vision}, 92\penalty0
  (1):\penalty0 1--31, 2011.

\bibitem[Cheng et~al.(2021)Cheng, Misra, Schwing, Kirillov, and
  Girdhar]{supp-cheng2021mask2former}
Bowen Cheng, Ishan Misra, Alexander~G. Schwing, Alexander Kirillov, and Rohit
  Girdhar.
\newblock Masked-attention mask transformer for universal image segmentation.
\newblock \emph{arXiv}, 2021.

\bibitem[Chollet(2017)]{supp-chollet2017xceptiondeeplearningdepthwise}
François Chollet.
\newblock Xception: Deep learning with depthwise separable convolutions, 2017.

\bibitem[Cordts et~al.(2016)Cordts, Omran, Ramos, Rehfeld, Enzweiler, Benenson,
  Franke, Roth, and Schiele]{supp-cordts2016cityscapesdatasetsemanticurban}
Marius Cordts, Mohamed Omran, Sebastian Ramos, Timo Rehfeld, Markus Enzweiler,
  Rodrigo Benenson, Uwe Franke, Stefan Roth, and Bernt Schiele.
\newblock The cityscapes dataset for semantic urban scene understanding, 2016.

\bibitem[Couairon et~al.(2025)Couairon, Chambon, Serrano, Haugeard, Cord, and
  Thome]{supp-couairon2025jafar}
Paul Couairon, Loick Chambon, Louis Serrano, Jean-Emmanuel Haugeard, Matthieu
  Cord, and Nicolas Thome.
\newblock Jafar: Jack up any feature at any resolution, 2025.

\bibitem[Deng et~al.(2009)Deng, Dong, Socher, Li, Li, and
  Fei-Fei]{supp-deng2009imagenet}
Jia Deng, Wei Dong, Richard Socher, Li-Jia Li, Kai Li, and Li Fei-Fei.
\newblock Imagenet: A large-scale hierarchical image database.
\newblock In \emph{2009 IEEE conference on computer vision and pattern
  recognition}, pages 248--255. Ieee, 2009.

\bibitem[Huang et~al.(2025)Huang, Chen, Havrylov, Geiger, and
  Zhang]{supp-huang2025loftup}
Haiwen Huang, Anpei Chen, Volodymyr Havrylov, Andreas Geiger, and Dan Zhang.
\newblock Loftup: Learning a coordinate-based feature upsampler for vision
  foundation models.
\newblock \emph{arXiv preprint arXiv:2504.14032}, 2025.

\bibitem[Jain et~al.(2022)Jain, Li, Chiu, Hassani, Orlov, and
  Shi]{supp-jain2022oneformertransformerruleuniversal}
Jitesh Jain, Jiachen Li, MangTik Chiu, Ali Hassani, Nikita Orlov, and Humphrey
  Shi.
\newblock Oneformer: One transformer to rule universal image segmentation,
  2022.

\bibitem[Lacoste et~al.(2019)Lacoste, Luccioni, Schmidt, and
  Dandres]{supp-lacoste2019quantifying}
Alexandre Lacoste, Alexandra Luccioni, Victor Schmidt, and Thomas Dandres.
\newblock Quantifying the carbon emissions of machine learning.
\newblock \emph{arXiv preprint arXiv:1910.09700}, 2019.

\bibitem[Lin et~al.(2015)Lin, Maire, Belongie, Bourdev, Girshick, Hays, Perona,
  Ramanan, Zitnick, and Dollár]{supp-lin2015microsoftcococommonobjects}
Tsung-Yi Lin, Michael Maire, Serge Belongie, Lubomir Bourdev, Ross Girshick,
  James Hays, Pietro Perona, Deva Ramanan, C.~Lawrence Zitnick, and Piotr
  Dollár.
\newblock Microsoft coco: Common objects in context, 2015.

\bibitem[Loshchilov and
  Hutter(2019)]{supp-loshchilov2019decoupledweightdecayregularization}
Ilya Loshchilov and Frank Hutter.
\newblock Decoupled weight decay regularization, 2019.

\bibitem[Oquab et~al.(2023)Oquab, Darcet, Moutakanni, Vo, Szafraniec, Khalidov,
  Fernandez, Haziza, Massa, El-Nouby, et~al.]{supp-oquab2023dinov2}
Maxime Oquab, Timoth{\'e}e Darcet, Th{\'e}o Moutakanni, Huy Vo, Marc
  Szafraniec, Vasil Khalidov, Pierre Fernandez, Daniel Haziza, Francisco Massa,
  Alaaeldin El-Nouby, et~al.
\newblock Dinov2: Learning robust visual features without supervision.
\newblock \emph{arXiv preprint arXiv:2304.07193}, 2023.

\bibitem[Roberts et~al.(2021)Roberts, Ramapuram, Ranjan, Kumar, Bautista,
  Paczan, Webb, and
  Susskind]{supp-roberts2021hypersimphotorealisticsyntheticdataset}
Mike Roberts, Jason Ramapuram, Anurag Ranjan, Atulit Kumar, Miguel~Angel
  Bautista, Nathan Paczan, Russ Webb, and Joshua~M. Susskind.
\newblock Hypersim: A photorealistic synthetic dataset for holistic indoor
  scene understanding, 2021.

\bibitem[Sim{\'e}oni et~al.(2025)Sim{\'e}oni, Vo, Seitzer, Baldassarre, Oquab,
  Jose, Khalidov, Szafraniec, Yi, Ramamonjisoa, Massa, Haziza, Wehrstedt, Wang,
  Darcet, Moutakanni, Sentana, Roberts, Vedaldi, Tolan, Brandt, Couprie,
  Mairal, J{\'e}gou, Labatut, and Bojanowski]{supp-simeoni2025dinov3}
Oriane Sim{\'e}oni, Huy~V. Vo, Maximilian Seitzer, Federico Baldassarre, Maxime
  Oquab, Cijo Jose, Vasil Khalidov, Marc Szafraniec, Seungeun Yi, Micha{\"e}l
  Ramamonjisoa, Francisco Massa, Daniel Haziza, Luca Wehrstedt, Jianyuan Wang,
  Timoth{\'e}e Darcet, Th{\'e}o Moutakanni, Leonel Sentana, Claire Roberts,
  Andrea Vedaldi, Jamie Tolan, John Brandt, Camille Couprie, Julien Mairal,
  Herv{\'e} J{\'e}gou, Patrick Labatut, and Piotr Bojanowski.
\newblock {DINOv3}, 2025.

\bibitem[Walmer et~al.(2026)Walmer, Suri, Aggarwal, and
  Shrivastava]{supp-walmer2026uplift}
Matthew Walmer, Saksham Suri, Anirud Aggarwal, and Abhinav Shrivastava.
\newblock Uplift: Efficient pixel-dense feature upsampling with local
  attenders.
\newblock \emph{arXiv preprint arXiv:2601.17950}, 2026.

\bibitem[Wimmer et~al.(2025)Wimmer, Truong, Rakotosaona, Oechsle, Tombari,
  Schiele, and Lenssen]{supp-wimmer2025anyup}
Thomas Wimmer, Prune Truong, Marie-Julie Rakotosaona, Michael Oechsle, Federico
  Tombari, Bernt Schiele, and Jan~Eric Lenssen.
\newblock Anyup: Universal feature upsampling.
\newblock \emph{arXiv preprint arXiv:2510.12764}, 2025.

\bibitem[Yang et~al.(2024)Yang, Kang, Huang, Xu, Feng, and
  Zhao]{supp-yang2024depthanythingunleashingpower}
Lihe Yang, Bingyi Kang, Zilong Huang, Xiaogang Xu, Jiashi Feng, and Hengshuang
  Zhao.
\newblock Depth anything: Unleashing the power of large-scale unlabeled data,
  2024.

\bibitem[Zhou et~al.(2022)Zhou, Loy, and Dai]{supp-zhou2022maskclip}
Chong Zhou, Chen~Change Loy, and Bo Dai.
\newblock Extract free dense labels from clip.
\newblock In \emph{European Conference on Computer Vision (ECCV)}, 2022.

\end{thebibliography}
    \bibliographystyle{ieeenat_fullname}
}

\clearpage
\setlength{\columnsep}{0.3125in}
\makeatletter
\let\maketitle\SavedMaketitle
\let\@maketitle\Saved@maketitle
\makeatother
\makeatletter
\gdef\@title{\ourmethod: Dense Semantic Features using Elastic Deformation - Supplementary Material}
\gdef\@author{}
\makeatother
\maketitle
\thispagestyle{empty}
\appendix

\setcounter{figure}{0}
\setcounter{table}{0}
\setcounter{equation}{0}
\renewcommand{\theHfigure}{supp.\arabic{figure}}
\renewcommand{\theHtable}{supp.\arabic{table}}
\renewcommand{\theHequation}{supp.\arabic{equation}}

\section{Implementation Details}


We train our model using batches drawn alternately from two datasets: Hypersim \citesupp{supp-roberts2021hypersimphotorealisticsyntheticdataset}, which provides ground truth for intrinsic decomposition and normal estimation, and ImageNet-22k \citesupp{supp-deng2009imagenet} for increased image diversity.

To avoid introducing additional datasets for semantic segmentation and depth prediction, we utilize pseudo-labels generated by pre-trained models, specifically, Mask2Former (COCO,ViT-Large) \citesupp{supp-cheng2021mask2former} and DepthAnything \citesupp{supp-yang2024depthanythingunleashingpower} respectively. These labels are computed for images across both of the training datasets.

%

%

To minimize parameter count and overhead on top of the base ViT prediction, we use depth-wise separable convolutions \citesupp{supp-chollet2017xceptiondeeplearningdepthwise}. We observe no performance hit compared to regular convolutions, with significantly faster training and inference, and vastly reduced memory usage.

We train in mixed-precision using an 8-bit version of the AdamW  optimizer \citesupp{supp-loshchilov2019decoupledweightdecayregularization}, with parameters $\epsilon=10^{-5}$ for stability, an effective batch size of 16 and an exponential learning rate decay starting at $10^{-3}$. The final model was trained on a single A100 GPU for 24 hours, amounting to 3.42kg of CO2-equivalent emissions, fully offset by the cloud provider (Google Cloud Services), and computed using the \href{https://mlco2.github.io/impact#compute}{MachineLearning Impact calculator} presented in \citesupp{supp-lacoste2019quantifying}.

\section{Further Quantitative Results}
\subsection{Open Vocabulary Evaluation}
We evaluate our model on zero-Shot open-vocabulary segmentation following the method presented in MaskCLIP \citesupp{supp-zhou2022maskclip}. This training-free approach computes pixel-wise predictions by taking the argmax over similarity scores between visual features and text embeddings generated from dataset labels in textual templates. Following JAFAR \citesupp{supp-couairon2025jafar}, we evaluate on  CLIP-ViT-B/16 and observe competitive results, demonstrating that \ourmethod preserves the original backbone features through its upsampling. Results against competitors that are compatible with CLIP-ViT-B/16 (JAFAR \citesupp{supp-couairon2025jafar}, AnyUp \citesupp{supp-wimmer2025anyup}, LoftUp \citesupp{supp-huang2025loftup}) are reported in Table \ref{tab:open_vocab}.

\begin{table}[ht]
    \centering
    \caption{Zero-Shot Open-Vocabulary segmentation on VOC and Ade20k datasets following the methodology used in JAFAR \citesupp{supp-couairon2025jafar} and introduced in MaskCLIP \citesupp{supp-zhou2022maskclip} paper. Competitive results indicating good alignment with original internal CLIP features.}
    \setlength{\tabcolsep}{3pt}
    \begin{tabular}{@{}l cc cc@{}}
        \toprule
        \multirow{2}{*}{\textbf{Method}} & \multicolumn{2}{c}{\textbf{VOC}} & \multicolumn{2}{c}{\textbf{ADE20K}} \\
        \cmidrule(lr){2-3} \cmidrule(lr){4-5}
        & mIoU($\uparrow$) & Acc($\uparrow$) & mIoU($\uparrow$) & Acc($\uparrow$) \\
        \midrule
        Bilinear & 51.25 & 68.18 & 9.05 & 25.15 \\
        JAFAR    & \textbf{63.99} & \textbf{77.91} & \uline{10.69} & {28.06} \\
        AnyUp & {57.59} & {73.40} & {10.71} & \textbf{28.95} \\
        LoftUp & {38.73} & {55.29} & {7.64} & {32.16} \\
        ElasticDino (ours) & \uline{58.43} & \uline{73.89} & \textbf{10.75} & \uline{28.28} \\
        \bottomrule
    \end{tabular}
    \label{tab:open_vocab}
\end{table}

\subsection{Positional Information Preservation}
While positional encoding related artifacts are alleviated, positional information still persists and can be extracted more easily from the features compared to competing upscaling methods. We train a simple pixel-wise head to predict normalized $(x,y)$ coordinates from upsampled features. Bilinear upsampling performs best, as expected for a linear target, but \ourmethod indeed preserves positional information. See accuracy in $[10^{-3}$ MSE$]$ results below.
\begin{table}[h]
\vspace{-5pt}
\centering
\setlength{\tabcolsep}{1.5pt}
\renewcommand{\arraystretch}{0.95}
\small
\begin{tabular}{lcccccc}
{} 
& JAFAR
& FeatUp
& AnyUp
& LoftUp
& Bilinear
& Ours \\
\midrule
$10^{-3}MSE$
& 5.3
& 11.5
& 4.2
& 64.0
& 3.2
& 4.5 \\
\end{tabular}
\label{tab:ed_ablation}
\end{table}
\vspace{-7pt}

\subsection{Semantic Segmentation and Depth Estimation over CLIP}
Following the comparison against other task agnostic upsampling methods (JAFAR \citesupp{supp-couairon2025jafar}, AnyUp \citesupp{supp-wimmer2025anyup}, LoftUp \citesupp{supp-huang2025loftup}, UPLift \citesupp{supp-walmer2026uplift}) over DINOv2 \citesupp{supp-oquab2023dinov2} and DINOv3 \citesupp{supp-simeoni2025dinov3}, we also compare recent works over CLIP \citesupp{supp-zhou2022maskclip} and get competitive results (Table \ref{tab:clip_table}).
We follow established evaluation protocols, as described by JAFAR \citesupp{supp-couairon2025jafar}, measuring mean Intersection over Union (mIoU) and pixel-level accuracy (Acc) for the semantic segmentation task over four datasets, and Root Mean Square Error (RMSE) and $\delta_1$ (measuring the  percentage of pixels with a relative error under 25\% of the ground-truth) for the depth estimation task over one dataset.

\begin{table*}
\centering
\footnotesize
\setlength{\tabcolsep}{4pt}

\caption{Comparison of models over CLIP backbone. Benchmarks are run using the method and code used for the JAFAR \cite{couairon2025jafar} paper. Bold = best, underline = second best.}
\label{tab:clip_table}
\begin{adjustbox}{width=\textwidth}
    \small 
    \begin{tabular}{l c c c c c c c c c c}
        \toprule
        \multirow{3}{*}{\textbf{CLIP}} & \multicolumn{8}{c}{\textbf{Semantic Segmentation}} & \multicolumn{2}{c}{\textbf{Depth Est.}} \\
        \cmidrule(lr){2-9} \cmidrule(lr){10-11}
        & \multicolumn{2}{c}{COCO} & \multicolumn{2}{c}{VOC} & \multicolumn{2}{c}{ADE20K} & \multicolumn{2}{c}{Cityscapes} & \multicolumn{2}{c}{COCO} \\
        \cmidrule(lr){2-3} \cmidrule(lr){4-5} \cmidrule(lr){6-7} \cmidrule(lr){8-9} \cmidrule(lr){10-11}
        & mIoU($\uparrow$) & Acc($\uparrow$) & mIoU($\uparrow$) & Acc($\uparrow$) & mIoU($\uparrow$) & Acc($\uparrow$) & mIoU($\uparrow$) & Acc($\uparrow$) & $\delta_1$($\uparrow$) & RMSE($\downarrow$) \\
        \midrule
        JAFAR \cite{couairon2025jafar} & \textbf{58.16} & \textbf{78.52} & \uline{79.07} & \textbf{94.69} &
        \textbf{36.65} & \textbf{71.39} & \textbf{50.98} & \textbf{91.23} & \uline{45.75} & \uline{0.92} \\
        AnyUp \cite{wimmer2025anyup} & 53.05 & 74.54 & 69.85 & 91.46 & 31.16 & 67.35 & 39.63 & 86.62 & 44.81 & 0.96 \\
        ED (ours) & \uline{57.86} & \uline{78.23} & \textbf{79.26} & \uline{94.67} & \uline{35.5} & \uline{71.22} & \uline{46.21} & \uline{90.67} & \textbf{48.20} & \textbf{0.86} \\
        \bottomrule
    \end{tabular}
\end{adjustbox}
\end{table*}

\subsection{Semantic Segmentation and Depth Estimation for no-seg and no-depth \ourmethod Variants}
We evaluated \ourmethod variants trained without the depth and the segmentation tasks in the same framework used to evaluate the full variant against task agnostic competitors over DINOv2 \citesupp{supp-oquab2023dinov2}. This is to strengthen the claim that \ourmethod only uses these tasks for structural guidance and does not rely on task specific information. Performance remains strong on these held-out tasks (Table \ref{tab:dino2_no_seg_no_depth_table}).

\begin{table*}[t]
\centering
\footnotesize
\setlength{\tabcolsep}{4pt}

\caption{Comparison of no-seg, no-depth \ourmethod variants against fully trained model. Benchmarks are run using the method and code used for the JAFAR\cite{couairon2025jafar} paper. Bold = best, underline = second best.}
\label{tab:dino2_no_seg_no_depth_table}

\begin{adjustbox}{width=\textwidth}
\begin{tabular}{l c c c c c c c c c c}
\toprule
        \multirow{3}{*}{\textbf{DINOv2}} & \multicolumn{8}{c}{\textbf{Semantic Segmentation}}
& \multicolumn{2}{c}{\begin{tabular}[c]{@{}c@{}}\textbf{Depth Est.}\end{tabular}} \\

\cmidrule(lr){2-9}
\cmidrule(lr){10-11}

& \multicolumn{2}{c}{COCO}
& \multicolumn{2}{c}{VOC}
& \multicolumn{2}{c}{ADE20K}
& \multicolumn{2}{c}{Cityscapes}
& \multicolumn{2}{c}{COCO} \\

\cmidrule(lr){2-9}
\cmidrule(lr){10-11}

Method
& {mIoU$\uparrow$} & {Acc$\uparrow$}
& {mIoU$\uparrow$} & {Acc$\uparrow$}
& {mIoU$\uparrow$} & {Acc$\uparrow$}
& {mIoU$\uparrow$} & {Acc$\uparrow$}
& {$\delta_1\uparrow$} & {RMSE$\downarrow$} \\

\cmidrule(lr){1-1}
\cmidrule(lr){2-3}
\cmidrule(lr){4-5}
\cmidrule(lr){6-7}
\cmidrule(lr){8-9}
\cmidrule(lr){10-11}











ED (no-depth variant)
& 61.7 & 81.1
& 84.1 & 96.3
& 41.3 & 75.5
& \uline{60.1} & 93.1
& 60.5 & 0.64 \\

ED (no-seg variant)
& \uline{62.5} & \uline{81.6}
& \uline{85.9} & \textbf{96.8}
& \uline{41.5} & \uline{75.8}
& 59.8 & \uline{93.3}
& \textbf{64.8} & \textbf{0.56} \\


ED (full variant)
& \textbf{62.72} & \textbf{81.74}
& \textbf{86.00} & \uline{96.78}
& \textbf{42.10} & \textbf{76.00}
& 61.33 & 93.44
& \uline{64.14} & \uline{0.57} \\

\bottomrule
\end{tabular}
\end{adjustbox}

\vspace{1em}

\end{table*}

\section{Further Qualitative Results}
\begin{figure*}
    \centering
    \makebox[\textwidth][c]{%
    \includegraphics[width=1\linewidth]{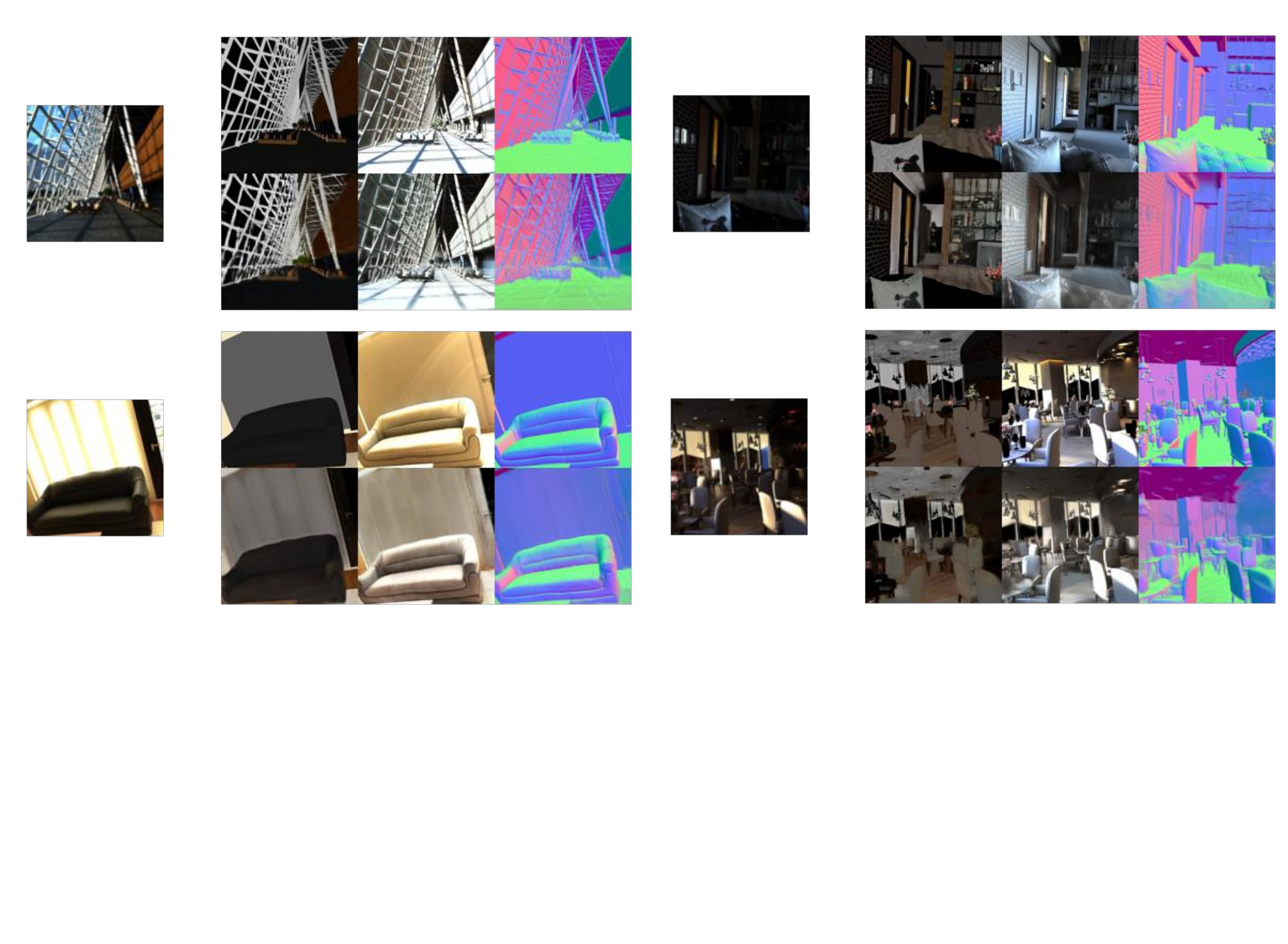}
    }
    \vspace{-1.3in}
    \caption{Intrinsic decomposition estimation trained on Hypersim \citesupp{supp-roberts2021hypersimphotorealisticsyntheticdataset} with a convolutional head at resolution 128x128. For each image, the decomposition is shown on the right: top row is the ground truth and bottom row is the prediction from ElasticDino. Left to right: albedo, diffuse shading, surface normals. Even in poor light conditions, we obtain strong qualitative results.}
    \label{fig:seg_city_transformers}
\end{figure*}

In this section we offer additional qualitative results. 
\paragraph{Dedicated Segmentation models.} We evaluate \ourmethod against two leading dedicated segmentation specific approaches Mask2Former \citesupp{supp-cheng2021mask2former} and OneFormer \citesupp{supp-jain2022oneformertransformerruleuniversal} over the Cityscapes\citesupp{supp-cordts2016cityscapesdatasetsemanticurban} dataset. We showcase that even though our approach is task agnostic we surpass leading dedicated segmentation models using only our lightweight operations over the DINOv2\citesupp{supp-oquab2023dinov2} backbone, see Figure \ref{fig:seg_city_transformers}. 

\begin{figure*}
    \centering
    \makebox[\textwidth][c]{%
    \includegraphics[width=1\linewidth]{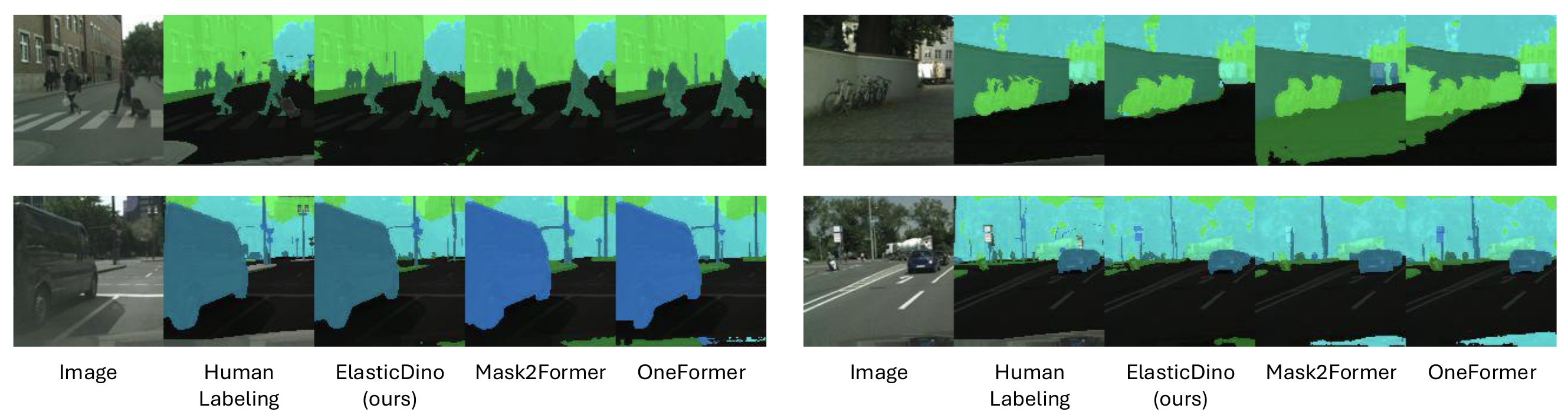}
    }
    \caption{Qualitative comparison of \ourmethod to dedicated segmentation approaches (Mask2Former~\citesupp{supp-cheng2021mask2former},OneFormer~\citesupp{supp-jain2022oneformertransformerruleuniversal}) over the cityScapes dataset.}
    \label{fig:intrinsic}
\end{figure*}

\begin{figure*}
    
    \includegraphics[trim={3cm 0 5cm 0 },clip,width=1\linewidth]{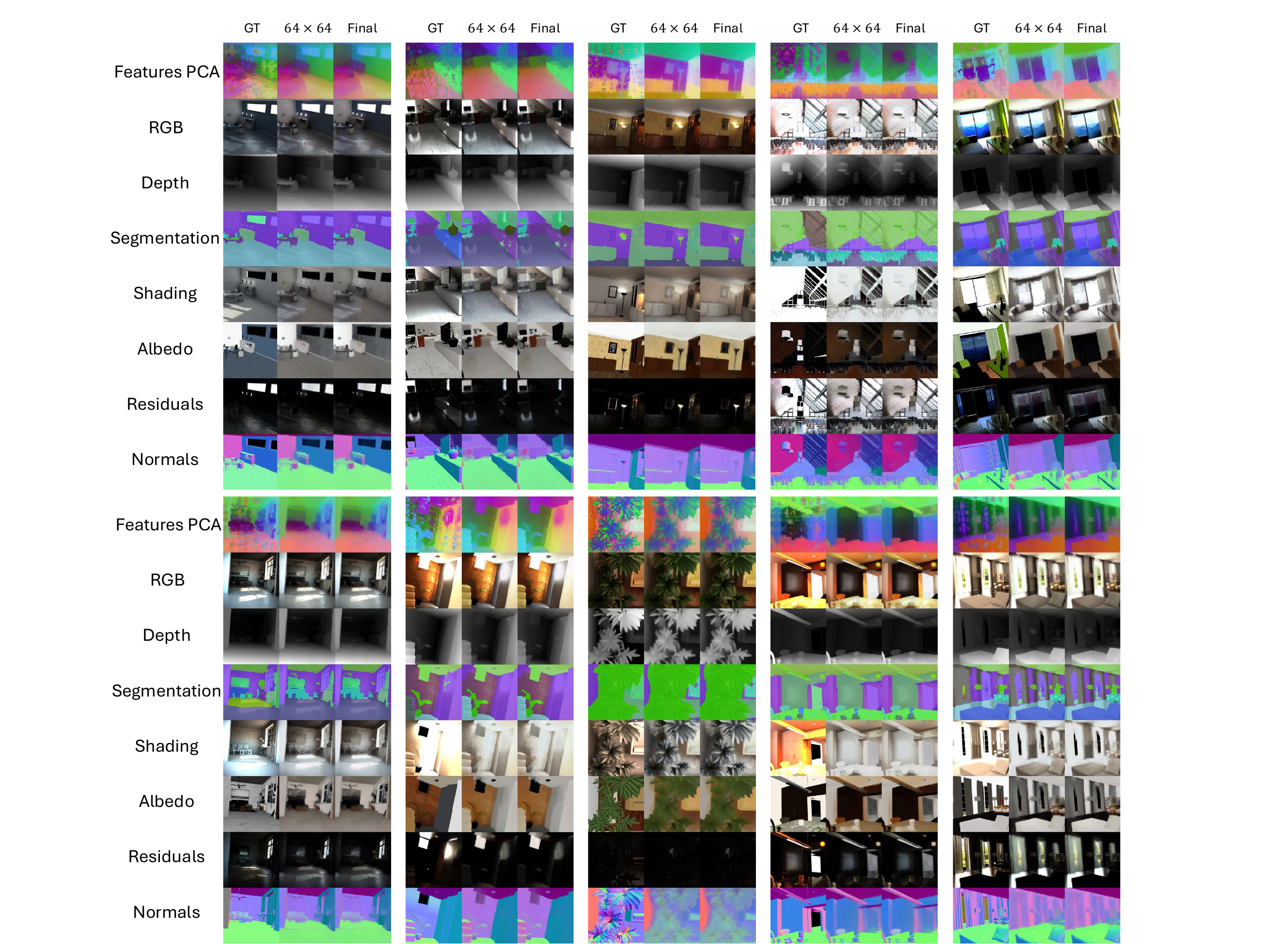}
    \centering
    \caption{Qualitative evaluation of our auxiliary training tasks. During training, our encoder is trained various tasks, each with an individual head and the minimal receptive field of one pixel. Their results are demonstrated here. Top to bottom: PCA of the upscaled features, RGB reconstruction, depth estimation, semantic segmentation, and four tasks of intrinsic decomposition (shading, albeo, residuals, normals). For each image, we show (left) its GT labels (or the features before upscaling for the PCA), intermediate results of our heads after upscaling to size $64\times64$ (middle), and the final results of size $256\times256$ (right). Zoomed-in viewing recommended. }
    \label{fig:heads}
\end{figure*}

\paragraph{Training heads.} \ourmethod employs different tasks to train the upscaler, by co-training a set of local and minimalistic $1\times1$ convolution heads, each performing a different dense task, considering pixels individually. In Figure \ref{fig:heads}, we demonstrate the results of these seven heads, both in their final step, where the features are of $256\times256$ resolution, and in an intermediate point in the network, where the resolution is upscaled from $32\times32$ to only $64\times64$. We further demonstrate qualitative results for intrinsic decomposition using the probe heads over the validation set of HyperSim \citesupp{supp-roberts2021hypersimphotorealisticsyntheticdataset} in Figure \ref{fig:intrinsic}.

\begin{figure*}
    \centering
    \includegraphics[trim={0 0 8cm 0},clip,width=1\linewidth]{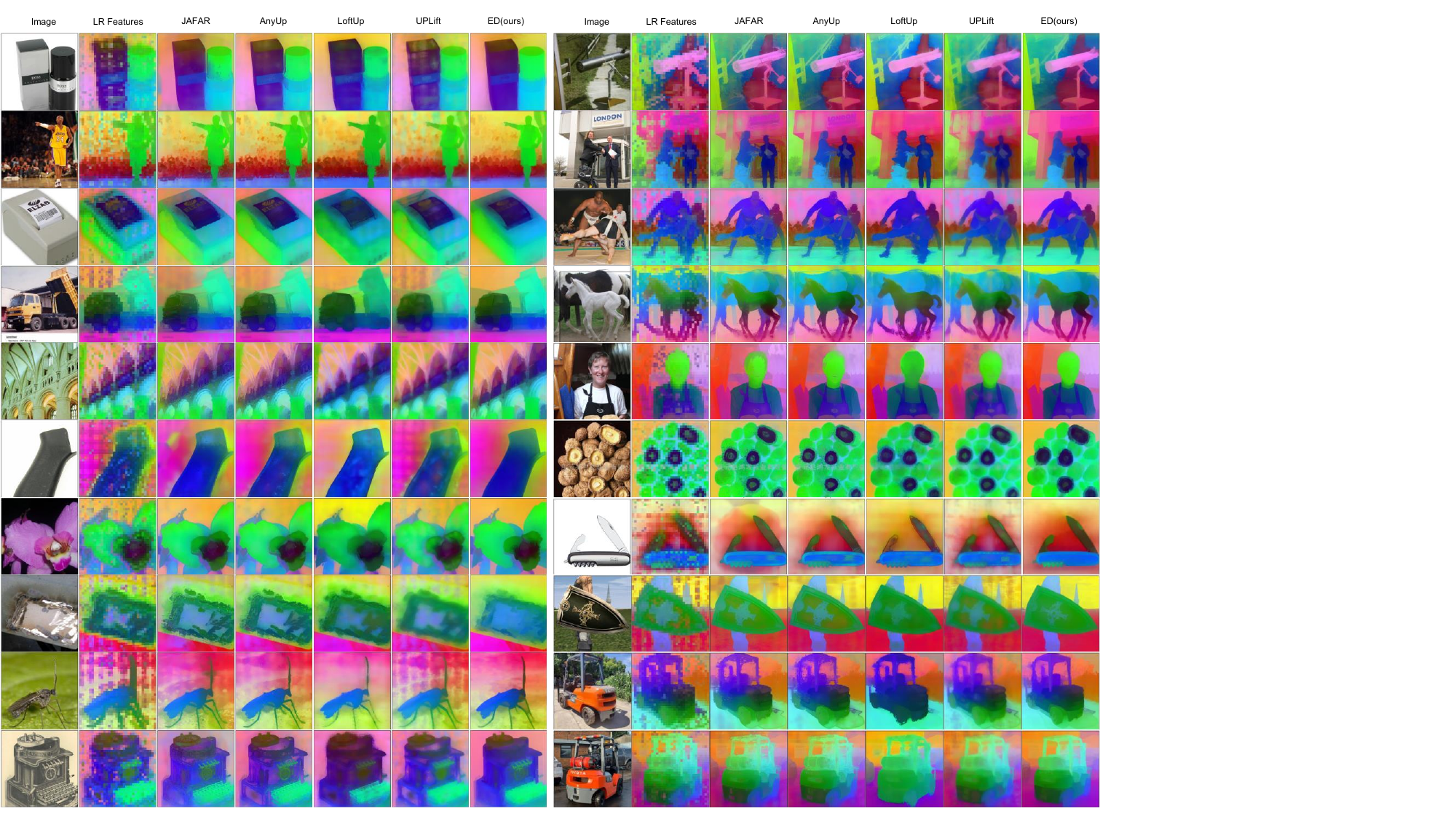}
    \caption{Additional qualitative comparisons through visualizations of the first three PCA components of the resulting features. From left to right: Input image, Original DINOv2 low-res features, JAFAR~\citesupp{supp-couairon2025jafar}, Anyup~\citesupp{supp-wimmer2025anyup}, LoftUp~\citesupp{supp-huang2025loftup}, UPLift~\citesupp{supp-walmer2026uplift}, ours. Zoomed-in viewing recommended. }
    \label{fig:pca_1}
\end{figure*}

\begin{figure*}
    \centering
    \includegraphics[trim={0 0 8cm 0},clip,width=1\linewidth]{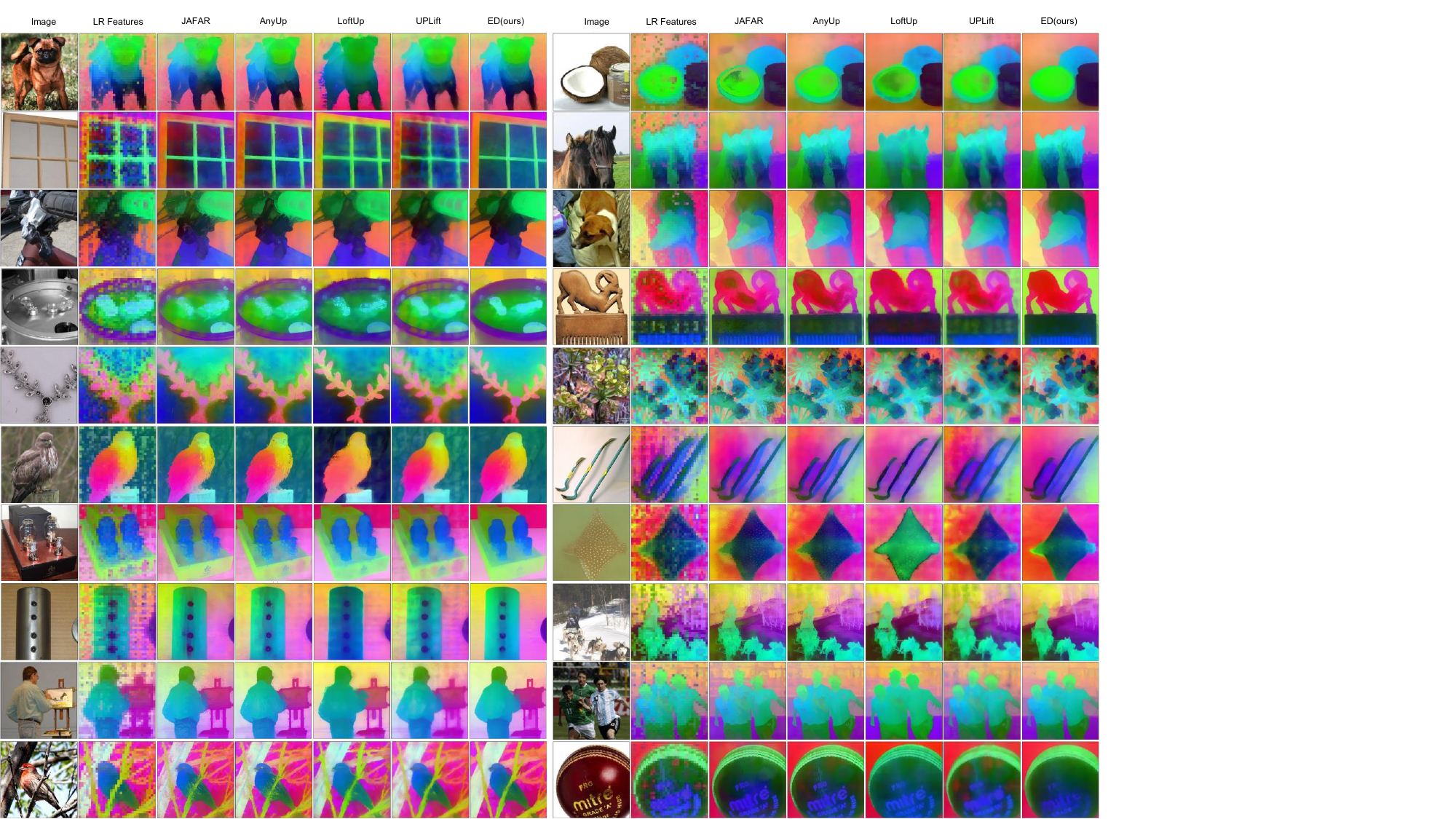}
    \caption{Additional qualitative comparisons through visualizations of the first three PCA components of the resulting features. From left to right: Input image, Original DINOv2 low-res features, JAFAR~\citesupp{supp-couairon2025jafar}, Anyup~\citesupp{supp-wimmer2025anyup}, LoftUp~\citesupp{supp-huang2025loftup}, UPLift~\citesupp{supp-walmer2026uplift}, ours. Zoomed-in viewing recommended.}
    \label{fig:pca_2}
\end{figure*}

\paragraph{PCA feature visualization.}   In the feature maps comparison figures (Figures \ref{fig:pca_1} and \ref{fig:pca_2}), we visualize the upscaled feature maps of \ourmethod using the first three components of a PCA decomposition and those of JAFAR \citesupp{supp-couairon2025jafar}, AnyUp \citesupp{supp-wimmer2025anyup}, LoftUp \citesupp{supp-huang2025loftup}, UPLift \citesupp{supp-walmer2026uplift}. As can be seen, our method achieves crisp features and cleaner maps from different artifacts as opposed to the competitors. It can be noticed that \ourmethod is better at removing artifacts originating from spatial-encodings and separates far and near objects (the bars of the forklift) as well as appearance from geometry (the shield).

\begin{figure*}
    \centering
    \includegraphics[trim={0 0 0 0},clip,width=2cm]{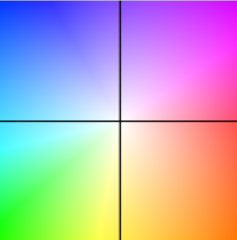}\\
    {\small Flow field color coding}\\[2mm]
    \includegraphics[trim={0 3cm 12.5cm 2cm},clip,width=1\linewidth]{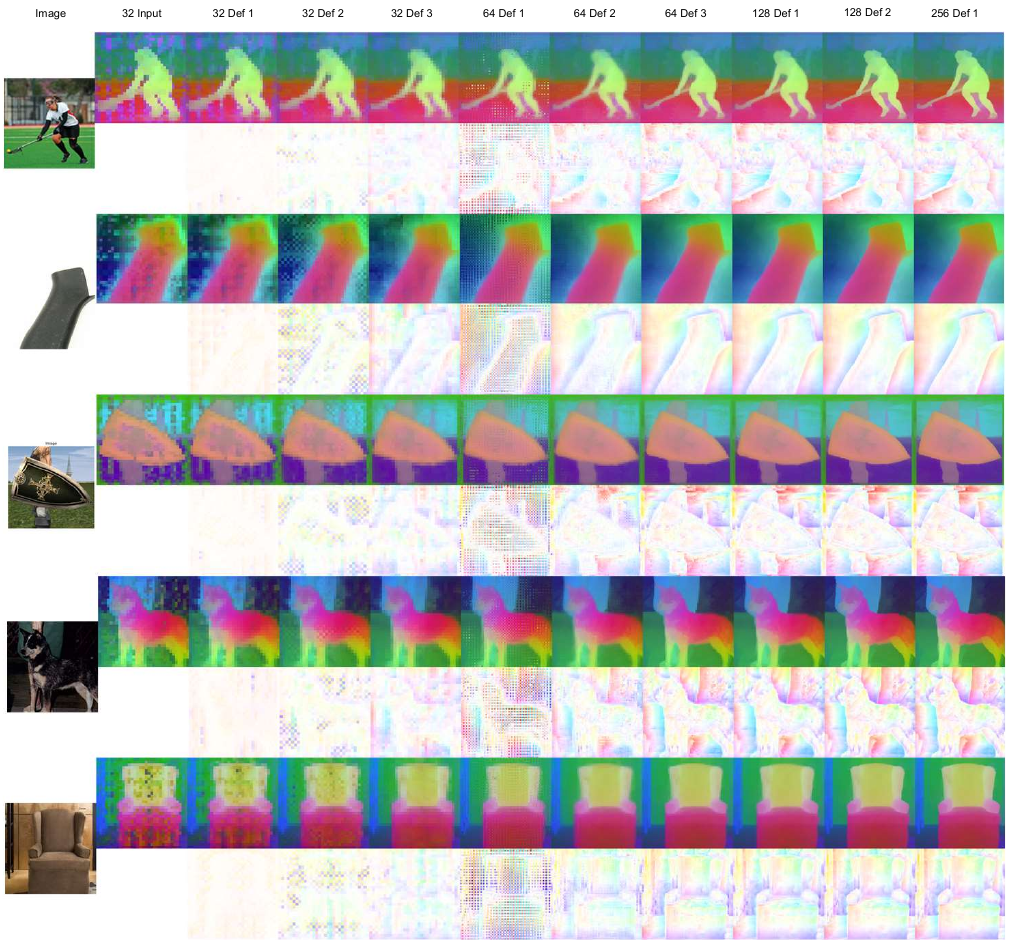}
    \caption{Visualization of deformations along the different resolution stages of the upscaling process and their effect on the features while being upscaled. Deformations are visualized using color wheel optical flow~\citesupp{supp-baker2011database}, showing where each pixel was sampled from. Features are visualized through the first three PCA components. For each image, top row shows the feature maps and bottom row shows the cumulative deformations so far. Zoomed-in viewing recommended. }
    \label{fig:deformations_vis_fig_1}
\end{figure*}

\begin{figure*}
    \centering
    \includegraphics[trim={0 0 0 0},clip,width=2cm]{figures/supp/WACV_optical_flow_color_coding.png}\\
    {\small Flow field color coding}\\[2mm]
    \includegraphics[trim={0 3cm 12.5cm 2cm},clip,width=1\linewidth]{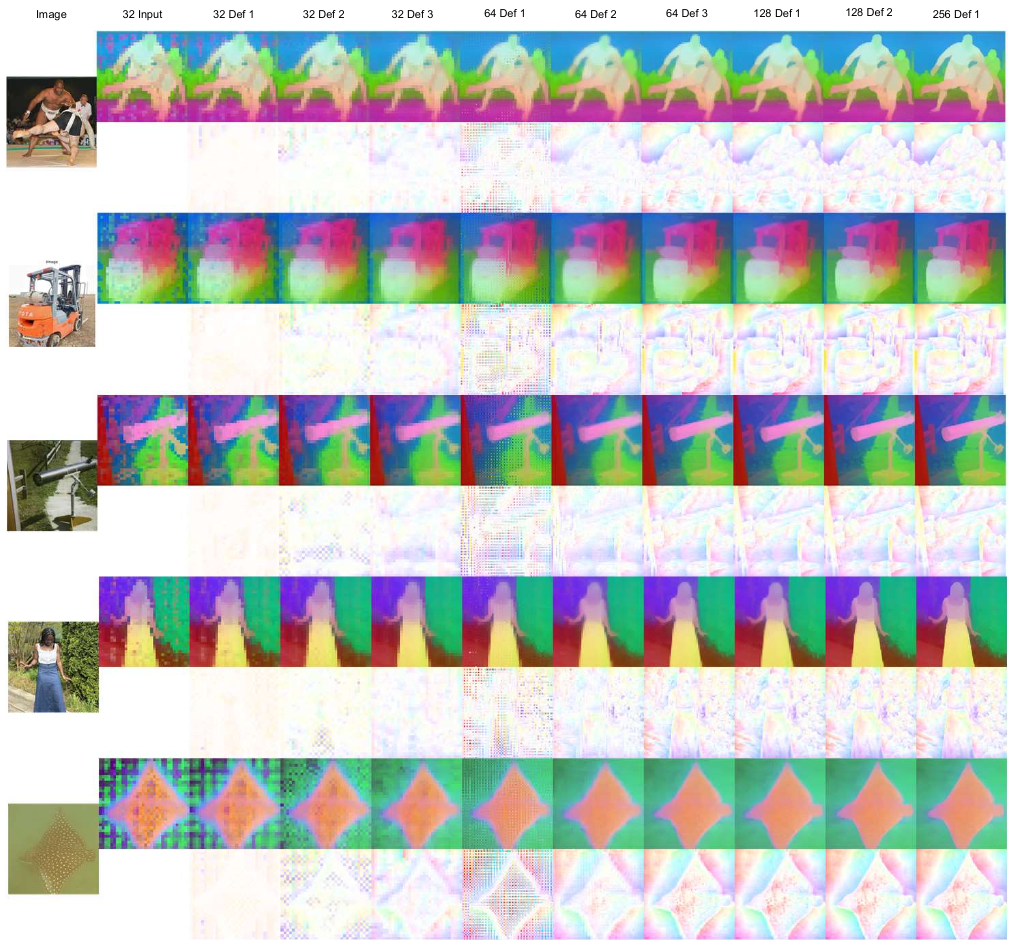}
    \caption{Visualization of deformations along the different resolution stages of the upscaling process and their effect on the features while being upscaled. Deformations are visualized using color wheel optical flow~\citesupp{supp-baker2011database}, showing where each pixel was sampled from. Features are visualized through the first three PCA components. For each image, top row shows the feature maps and bottom row shows the cumulative deformations so far. Zoomed-in viewing recommended. }
    \label{fig:deformations_vis_fig_2}
\end{figure*}

\paragraph{Visualization of deformations along upscaling process.} Figures \ref{fig:deformations_vis_fig_1}, \ref{fig:deformations_vis_fig_2} visualize the cumulative deformations through the process of feature upscaling for different images. Deformations are visualized using color wheel optical flow~\citesupp{supp-baker2011database} and features are visualized through the first three PCA components. It can be seen that deformations help to make the edges more crisp and reduce artifacts in the feature maps.

\begin{figure*}
    \centering
    \includegraphics[width=1\linewidth]{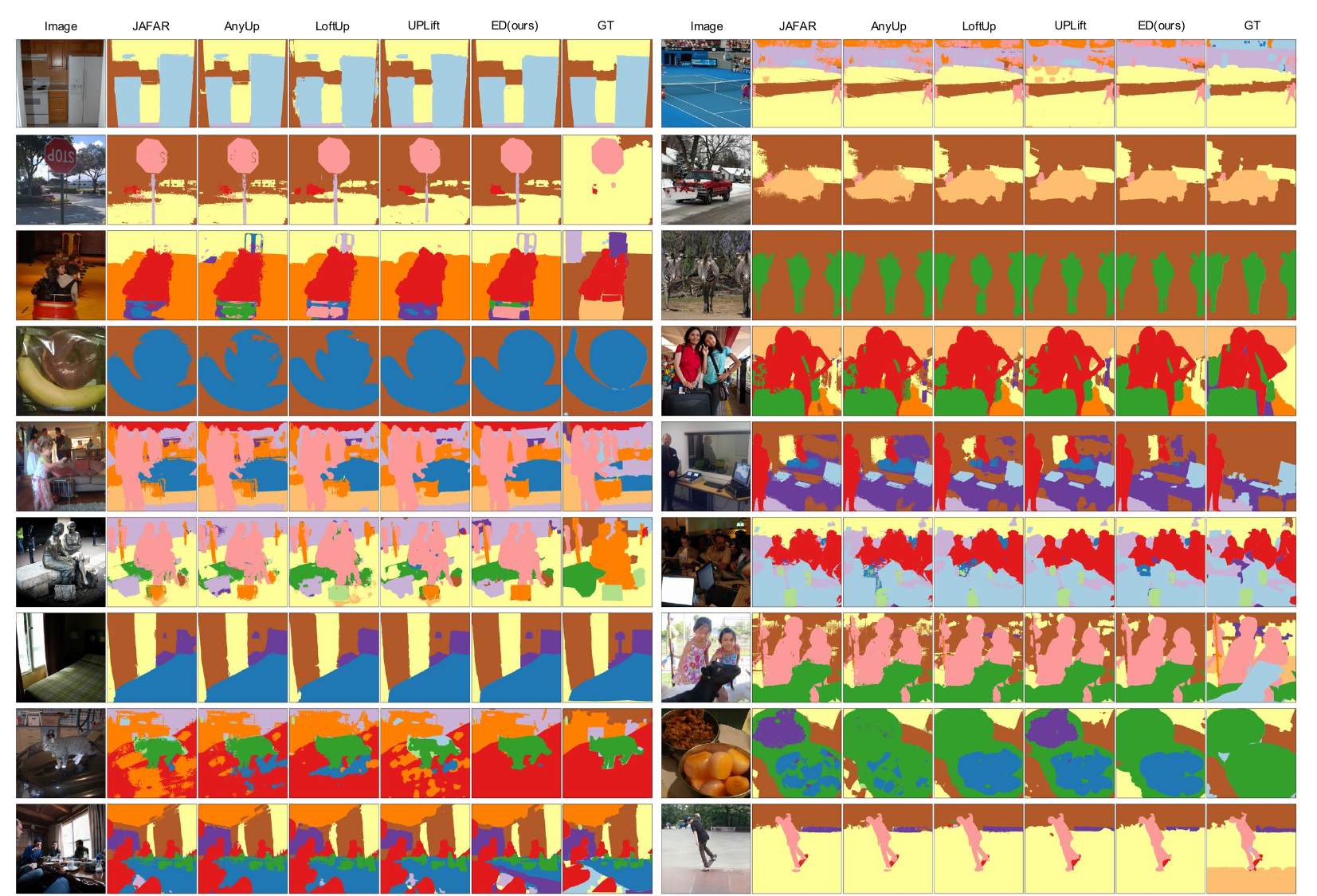}
    \caption{Additional qualitative comparisons for the segmentation task over the COCO dataset. From left to right: Input image, JAFAR~\citesupp{supp-couairon2025jafar}, Anyup~\citesupp{supp-wimmer2025anyup}, LoftUp~\citesupp{supp-huang2025loftup},
    UPLift~\citesupp{supp-walmer2026uplift}, ours, Ground Truth. Zoomed-in viewing recommended.}
    \label{fig:seg_coco}
\end{figure*}

\begin{figure*}
    \centering
    \includegraphics[width=1\linewidth]{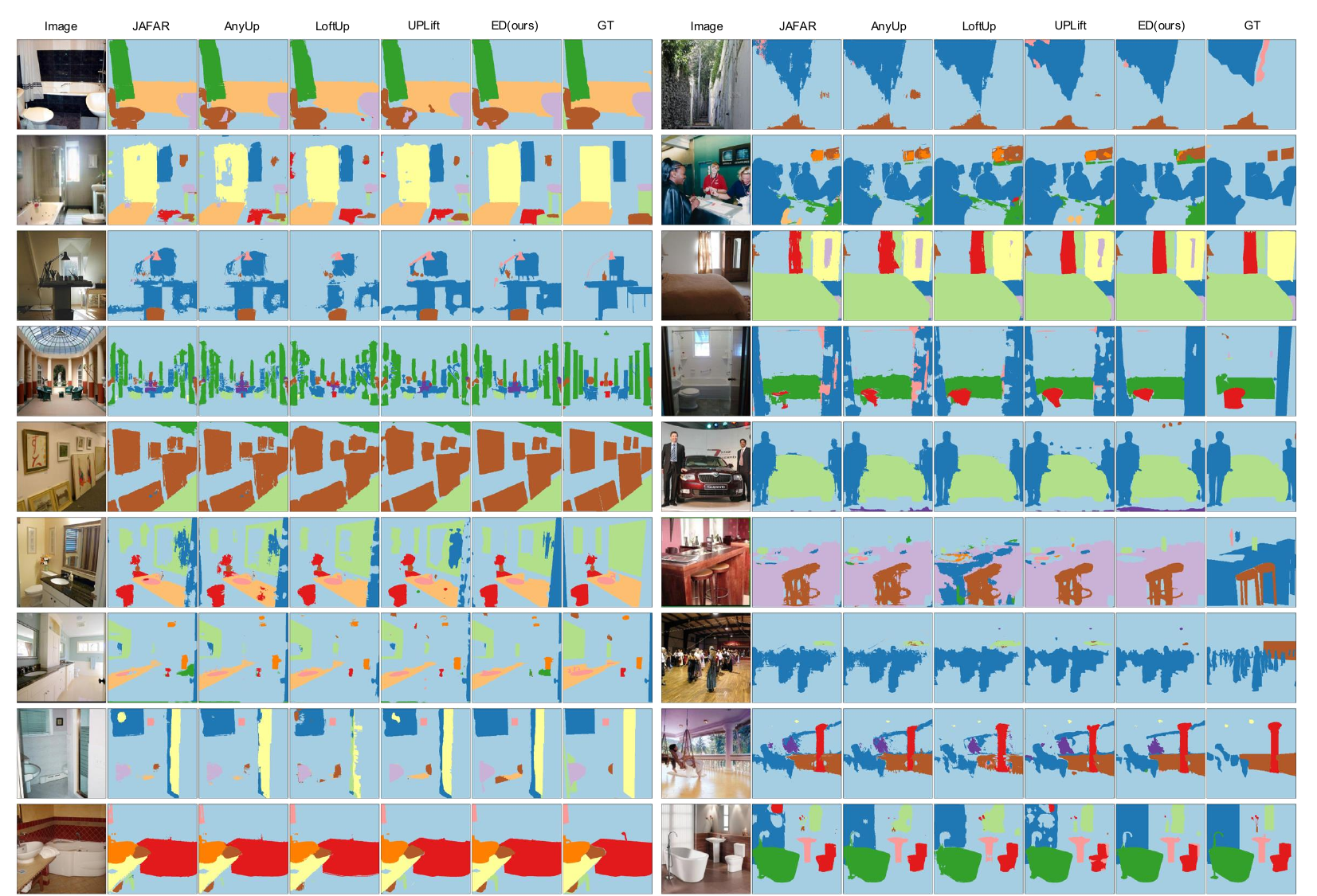}
    \caption{Additional qualitative comparisons for the segmentation task over the ADE20K dataset. From left to right: Input image, JAFAR~\citesupp{supp-couairon2025jafar}, Anyup~\citesupp{supp-wimmer2025anyup}, LoftUp~\citesupp{supp-huang2025loftup},
    UPLift~\citesupp{supp-walmer2026uplift}, ours, Ground Truth. Zoomed-in viewing recommended.}
    \label{fig:seg_ade}
\end{figure*}

\begin{figure*}
    \centering
    \includegraphics[width=1\linewidth]{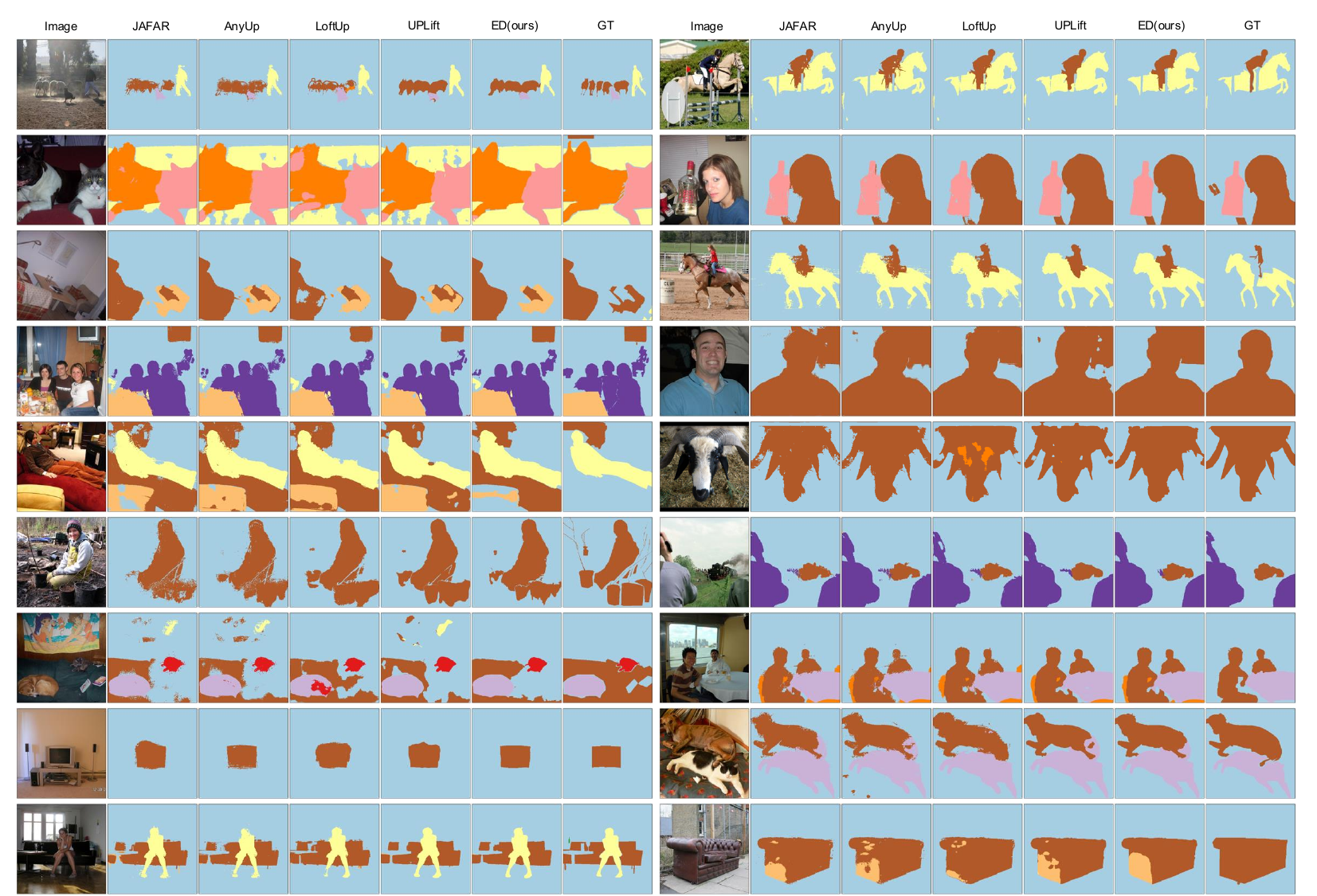}
    \caption{Additional qualitative comparisons for the segmentation task over the VOC dataset. From left to right: Input image, JAFAR~\citesupp{supp-couairon2025jafar}, Anyup~\citesupp{supp-wimmer2025anyup}, LoftUp~\citesupp{supp-huang2025loftup},
    UPLift~\citesupp{supp-walmer2026uplift}, ours, Ground Truth. Zoomed-in viewing recommended.}
    \label{fig:seg_voc}
\end{figure*}

\begin{figure*}
    \centering
    \includegraphics[width=1\linewidth]{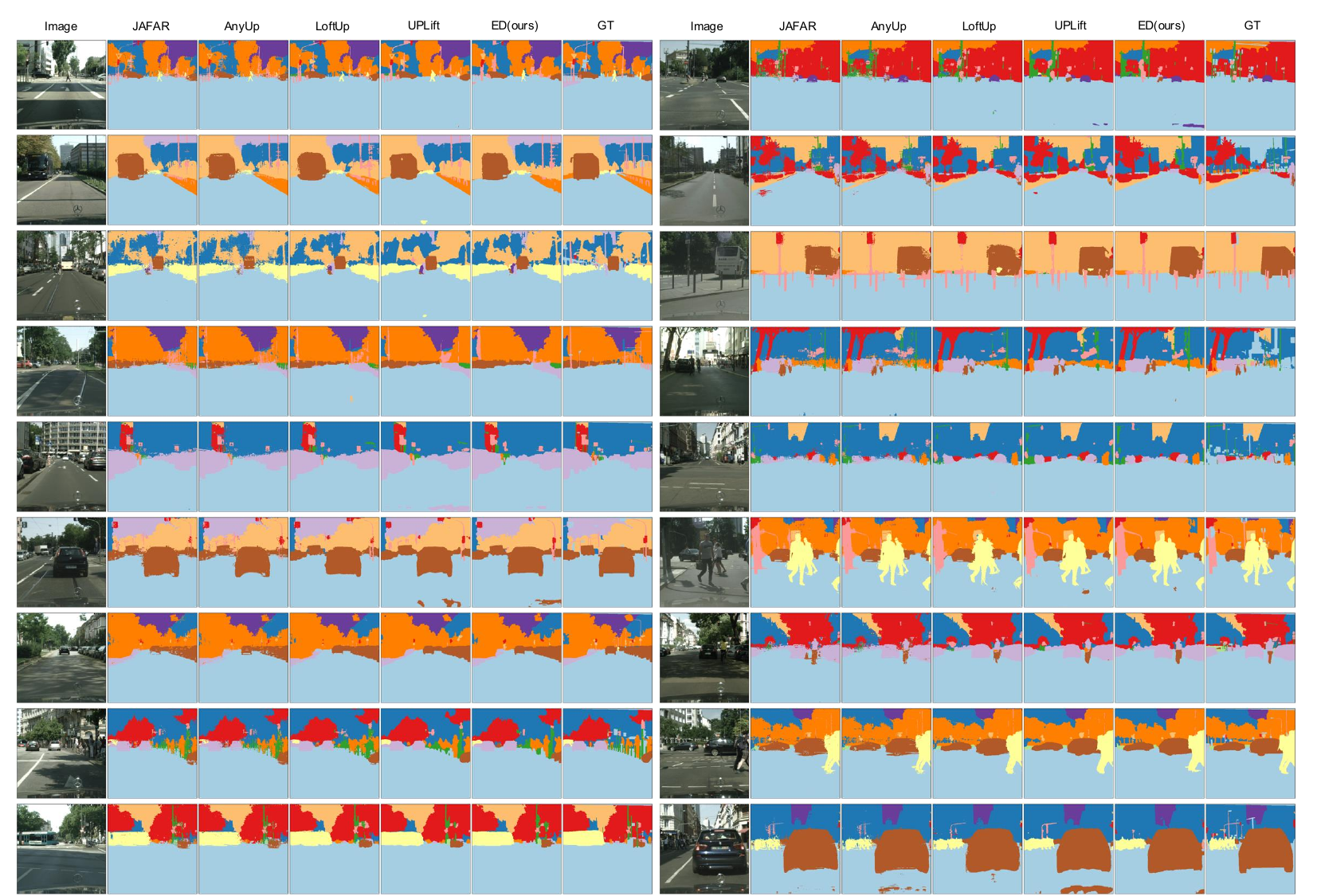}
    \caption{Additional qualitative comparisons for the segmentation task over the CityScapes dataset. From left to right: Input image, JAFAR~\citesupp{supp-couairon2025jafar}, Anyup~\citesupp{supp-wimmer2025anyup}, LoftUp~\citesupp{supp-huang2025loftup},
    UPLift ~\citesupp{supp-walmer2026uplift}, ours, Ground Truth. Zoomed-in viewing recommended.}
    \label{fig:seg_city}
\end{figure*}

\begin{figure*}
    \centering
    \includegraphics[trim={0 0 8cm 0},clip,width=1\linewidth]{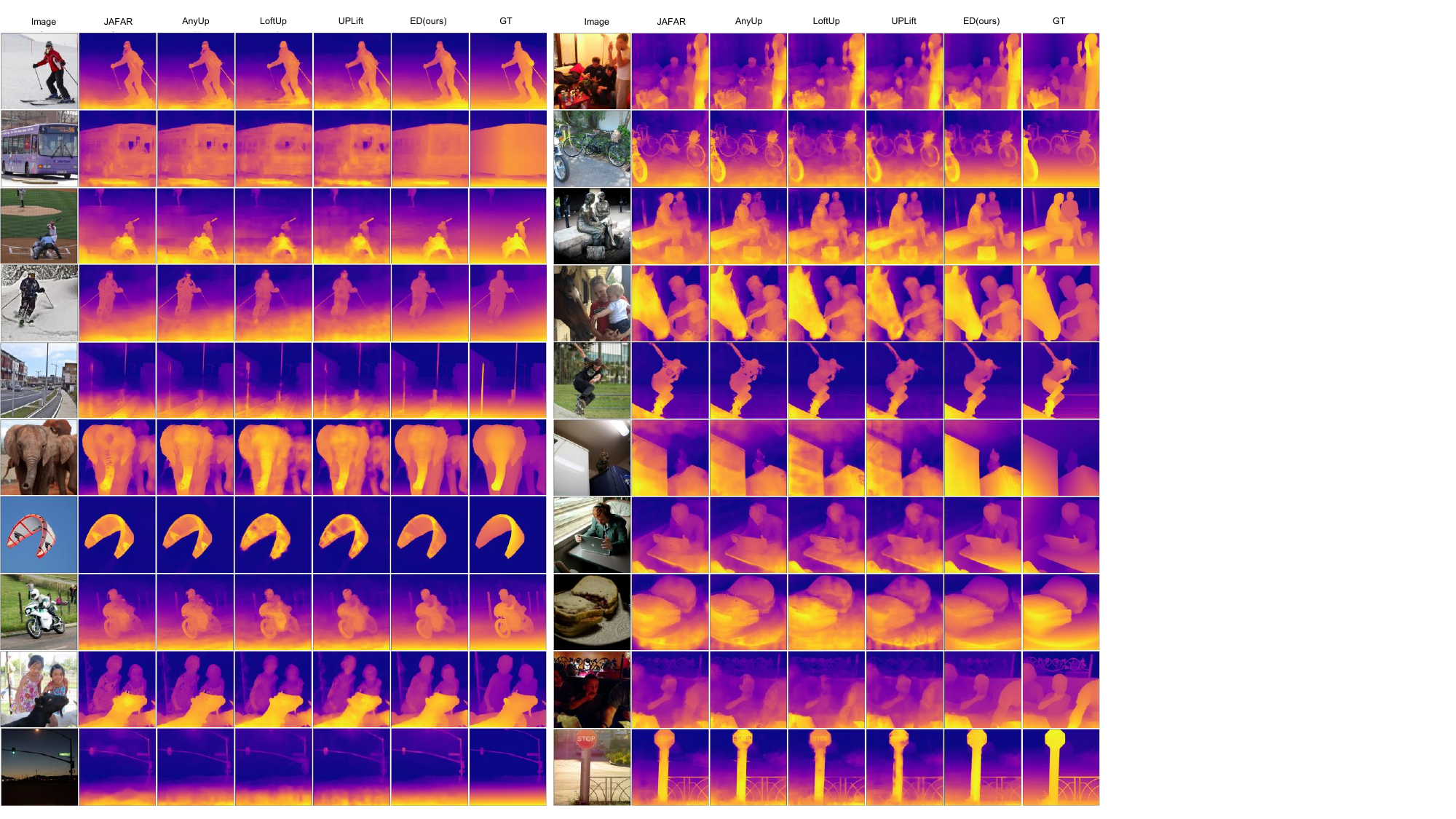}
    \caption{Additional qualitative comparisons for the depth estimation task over the COCO dataset. From left to right: Input image, JAFAR~\citesupp{supp-couairon2025jafar}, Anyup~\citesupp{supp-wimmer2025anyup}, LoftUp~\citesupp{supp-huang2025loftup},
    UPLift~\citesupp{supp-walmer2026uplift}, ours, Ground Truth. Zoomed-in viewing recommended.}
    \label{fig:depth_coco}
\end{figure*}

\paragraph{Linear segmentation and depth probe heads.} Finally, we also compare \ourmethod against JAFAR \citesupp{supp-couairon2025jafar}, AnyUp \citesupp{supp-wimmer2025anyup}, LoftUp \citesupp{supp-huang2025loftup} in the framework suggested by JAFAR where linear probe heads are trained over the upsamplers for segmentation and depth tasks displaying the information encoded inside the upsampled features produced by the model allows achieving top results for dense tasks such as semantic segmentation (Figures \ref{fig:seg_ade}, \ref{fig:seg_city}, \ref{fig:seg_coco}, \ref{fig:seg_voc}) and depth estimation (Figure \ref{fig:depth_coco}). It can be seen that \ourmethod achieves superior results with cleaner edges and often detects details other models are not able to. 
\section{Experiments to understand degraded performance over Cityscapes}
\begin{figure*}
    \centering
    \includegraphics[width=\columnwidth]{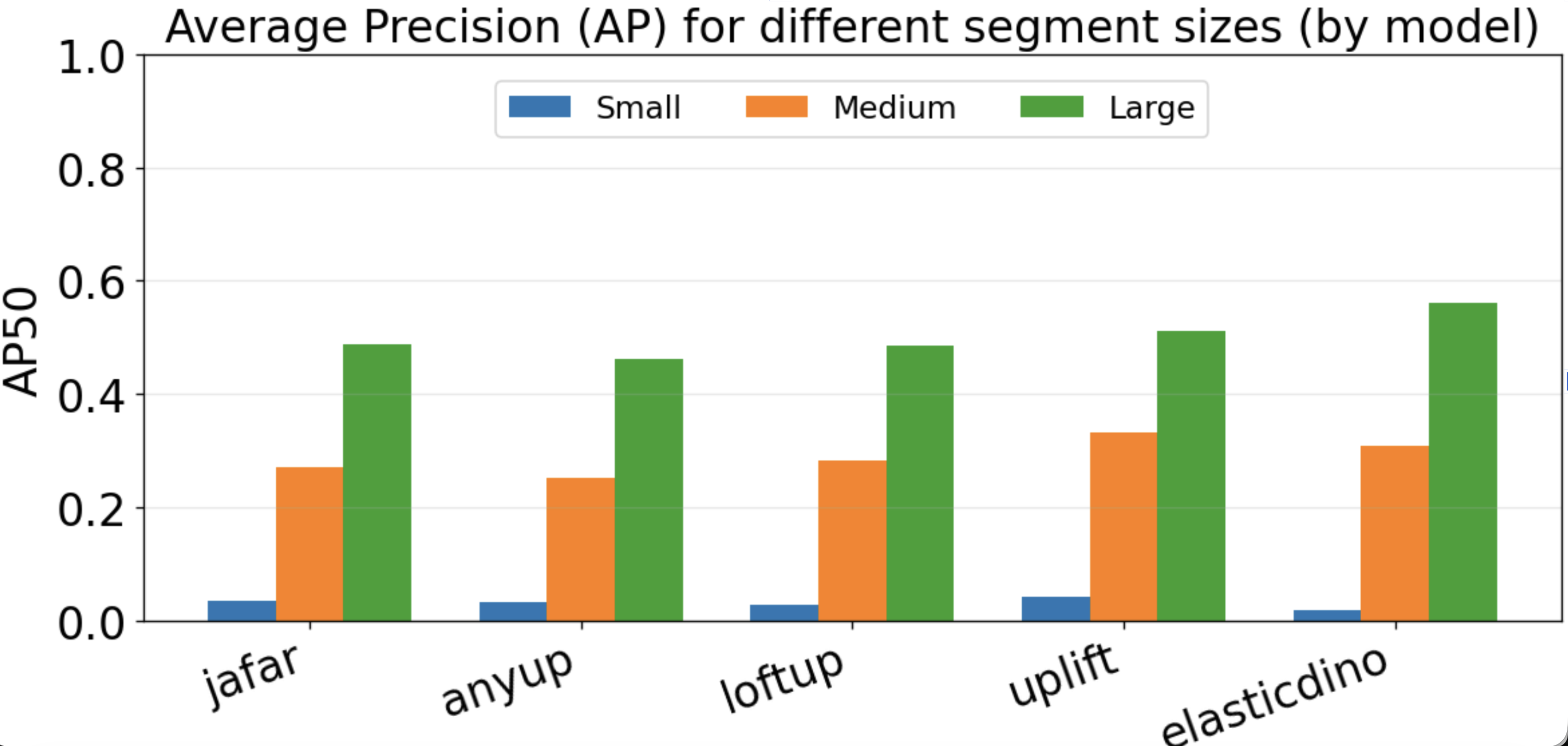}
    \includegraphics[width=\columnwidth]{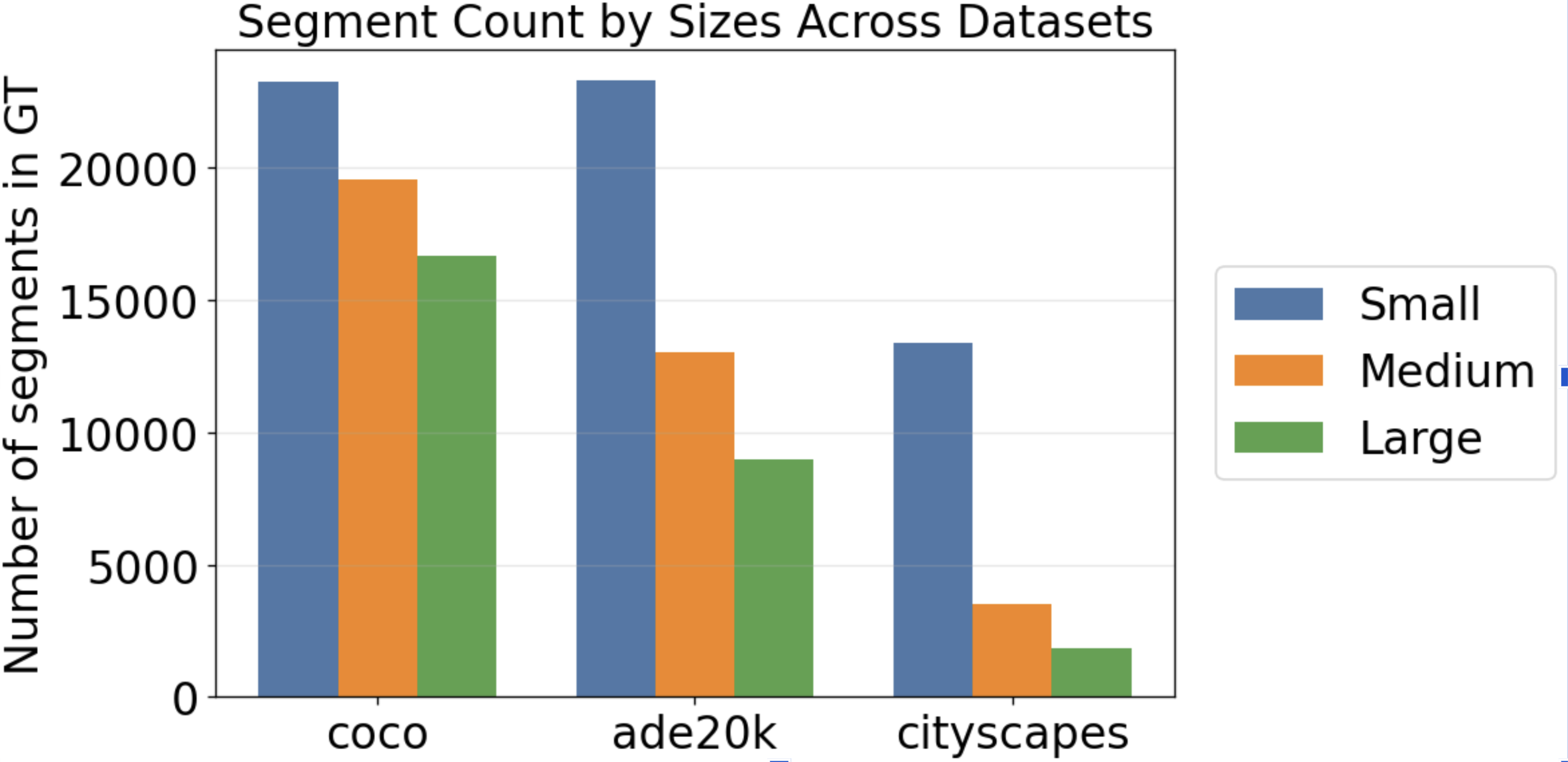}
    \caption{Left: Accuracy vs size. We perform better for larger segments. Right: sizes across datasets. Cityscapes consists of smaller objects.}
    \label{fig:rebut_city}
\end{figure*}

We investigated our change in relative performance on Cityscapes~\citesupp{supp-cordts2016cityscapesdatasetsemanticurban}, in contrast to all other datasets. We hypothesize it is because Cityscapes consists mainly of smaller objects and segments, connecting to our known limitation of filling "holes" in feature maps. To further investigate this we evaluated AP (Average Precision) metric for the different models over segments of different size categories Small (area $< 32^2$), Medium ($32^2 < $ area $ < 96^2$), Large (area $> 96^2$) as taken from COCO \citesupp{supp-lin2015microsoftcococommonobjects} instance segmentation evaluation methodology (see Figure \ref{fig:rebut_city}). We also measured distribution of different segment size categories over the COCO\citesupp{supp-lin2015microsoftcococommonobjects}, Cityscapes\citesupp{supp-cordts2016cityscapesdatasetsemanticurban}, Ade20k datasets.
From this can see that the lower performance of \ourmethod can over Cityscapes can be explained by its lower performance on small thin objects present in this dataset in particular.

\clearpage
{
    \small
    \bibliographystylesupp{ieeenat_fullname}
    \bibliographysupp{supplement-references}
}

\end{document}